\pdfoutput=1

\documentclass[11pt]{article}

\usepackage{acl}

\usepackage{times}
\usepackage{latexsym}

\usepackage[utf8]{inputenc} %
\usepackage[T1]{fontenc}    %
\usepackage{hyperref}       %
\usepackage{url}            %
\usepackage{booktabs}       %
\usepackage{amsfonts}       %
\usepackage{nicefrac}       %
\usepackage{microtype}      %
\usepackage{xcolor}         %
\usepackage{standalone}
\usepackage{latexsym}
\usepackage{amsmath}
\usepackage{amssymb}
\usepackage{amsthm}
\usepackage{graphicx}
\usepackage{subcaption}
\usepackage{array}
\usepackage{tabu}
\usepackage{makecell}
\usepackage{paralist}
\usepackage{cases}
\usepackage{diagbox}
\usepackage{enumitem}
\usepackage{soul}
\usepackage{multirow}
\usepackage{verbatim}
\usepackage{tabulary}
\usepackage{booktabs}
\usepackage[mathscr]{euscript}
\usepackage{mathtools}
\usepackage{algorithm}
\usepackage{algpseudocode}
\usepackage{stmaryrd}
\usepackage{tikz-dependency}
\usetikzlibrary{automata,decorations.markings,arrows,positioning,matrix,calc,patterns,angles,quotes,calc}
\usepackage{adjustbox}
\usepackage{tabularx}
\usepackage{xspace}
\usepackage{tabulary}
\usepackage{afterpage}
\usepackage{hyperref}
\usepackage{url}
\usepackage{bm}
\usepackage{color}
\usepackage{graphicx}
\usepackage{slashbox}
\usepackage[toc,page]{appendix}
\usepackage{makecell}
\usepackage{boldline}

\usepackage{bbm}
\usepackage{adjustbox}
\usepackage{tabu}
\usepackage{CJKutf8}
\usepackage{tablefootnote}

\definecolor{orange}{rgb}{1,0.5,0}
\definecolor{mdgreen}{rgb}{0.05,0.6,0.05}
\definecolor{mdblue}{rgb}{0,0,0.7}
\definecolor{dkblue}{rgb}{0,0,0.5}
\definecolor{dkgray}{rgb}{0.3,0.3,0.3}
\definecolor{slate}{rgb}{0.25,0.25,0.4}
\definecolor{gray}{rgb}{0.5,0.5,0.5}
\definecolor{ltgray}{rgb}{0.7,0.7,0.7}
\definecolor{purple}{rgb}{0.7,0,1.0}
\definecolor{lavender}{rgb}{0.65,0.55,1.0}

\definecolor{mypurple}{RGB}{111,61,121}
\definecolor{myblue}{RGB}{46,88,180}
\definecolor{myred}{RGB}{181,68,106}
\definecolor{myyellow}{RGB}{204,143,55}
\definecolor{lightgray}{gray}{0.9}
\definecolor{magenta}{HTML}{F3DFF1}
\definecolor{red}{HTML}{FF0000}
\definecolor{blue}{HTML}{0000FF}
\definecolor{darkgreen}{HTML}{228B22}

\newcommand{\textred}[1]{\textcolor{red}{#1}}
\newcommand{\textblue}[1]{\textcolor{blue}{#1}}

\newcommand{\interalia}[1]{\citep[\emph{inter alia}]{#1}}

\DeclareSymbolFont{extraup}{U}{zavm}{m}{n}
\DeclareMathSymbol{\vardiamond}{\mathalpha}{extraup}{87}

\newcolumntype{L}[1]{>{\raggedright\let\newline\\\arraybackslash\hspace{0pt}}m{#1}}
\newcolumntype{C}[1]{>{\centering\let\newline\\\arraybackslash\hspace{0pt}}m{#1}}
\newcolumntype{R}[1]{>{\raggedleft\let\newline\\\arraybackslash\hspace{0pt}}m{#1}}

\theoremstyle{definition}

\theoremstyle{remark}

\algrenewcommand{\algorithmiccomment}[1]{\leavevmode$\triangleright$ #1}

\setul{1pt}{.4pt}

\usepackage[shortcuts]{extdash}  %

\usepackage{blindtext}
\usepackage{graphicx}
\usepackage{capt-of}
\usepackage{booktabs}
\usepackage{varwidth}
\newsavebox\tmpbox

\usepackage{amsmath,amsfonts,bm}

\def\eqref#1{equation~\ref{#1}}

\def\1{\bm{1}}

\DeclareMathAlphabet{\mathsfit}{\encodingdefault}{\sfdefault}{m}{sl}
\SetMathAlphabet{\mathsfit}{bold}{\encodingdefault}{\sfdefault}{bx}{n}

\DeclareMathOperator*{\argmin}{arg\,min}

\newcommand{\bilboard}{\textsc{Billboard}\xspace}
\newcommand{\bilboards}{\textsc{Billboard}s\xspace}

\newcommand{\resolved}[1]{}

\newcommand{\genie}{\textsc{Genie}\xspace}
\newcommand{\thumb}{\textsc{THumB}\xspace}

\usepackage{pifont}
\newcommand{\xmark}{\textcolor{red}{\ding{55}}}
\newcommand{\cmark}{\textcolor{darkgreen}{\ding{51}}}

\newcommand{\com}[1]{}
\newcommand{\thumbup}[0]{\raisebox{-.2\height}{\includegraphics[width=.02\textwidth]{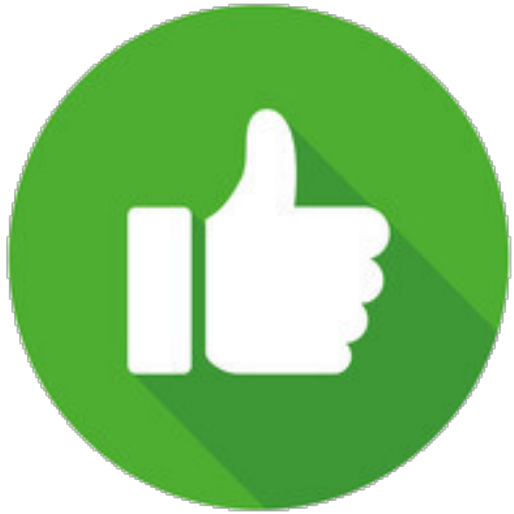}}}
\newcommand{\thumbdown}[0]{\raisebox{-.2\height}{\includegraphics[width=.02\textwidth]{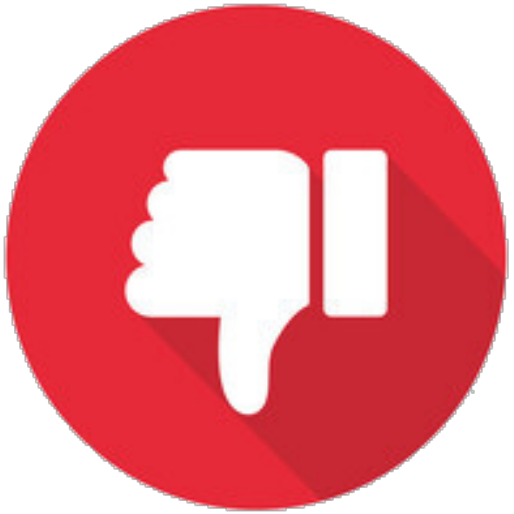}}}
\newcommand{\orangesquare}[0]{\raisebox{-.1\height}{\includegraphics[width=.02\textwidth]{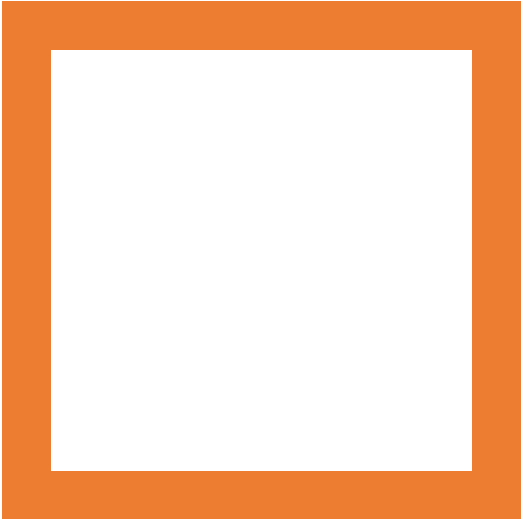}}}
\newcommand{\handshake}[0]{\raisebox{-.1\height}{\includegraphics[width=.025\textwidth]{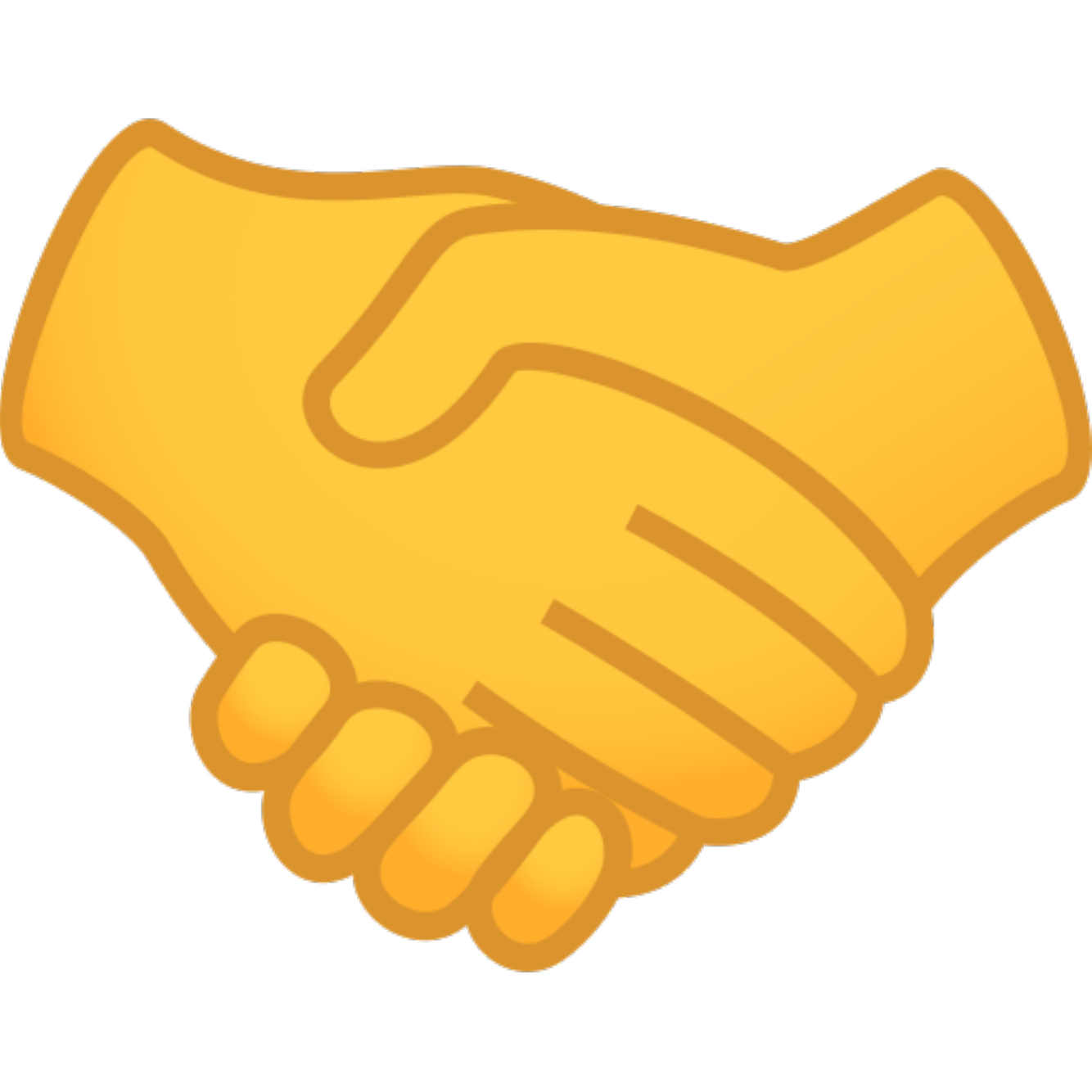}}}
\newcommand{\heart}[0]{\raisebox{-.1\height}{\includegraphics[width=.025\textwidth]{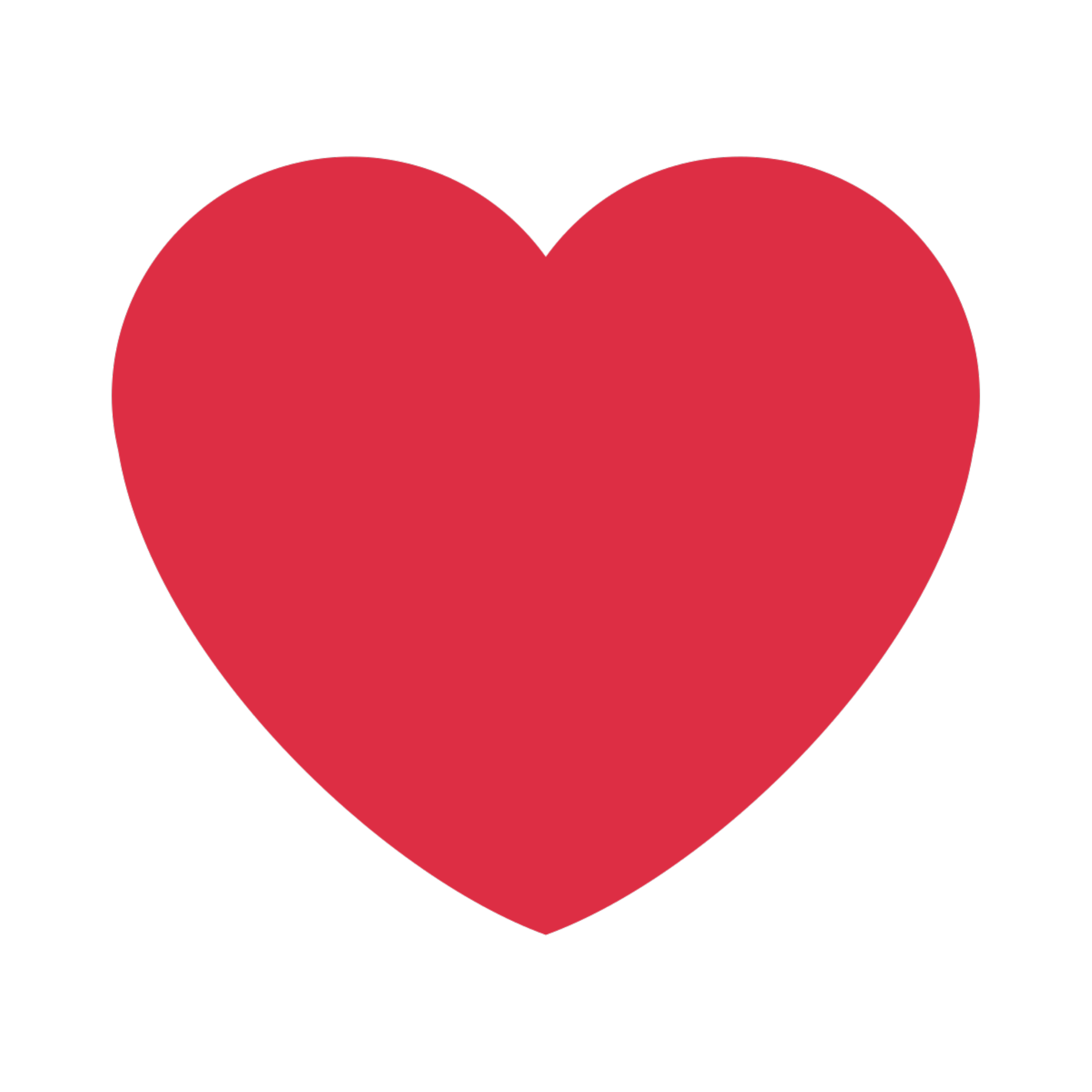}}}
\newcommand{\diamondemoji}[0]{\raisebox{-.0\height}{\includegraphics[width=.020\textwidth]{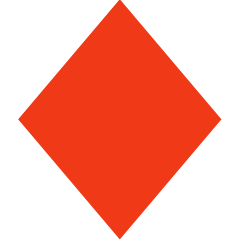}}}
\newcommand{\club}[0]{\raisebox{-.1\height}{\includegraphics[width=.025\textwidth]{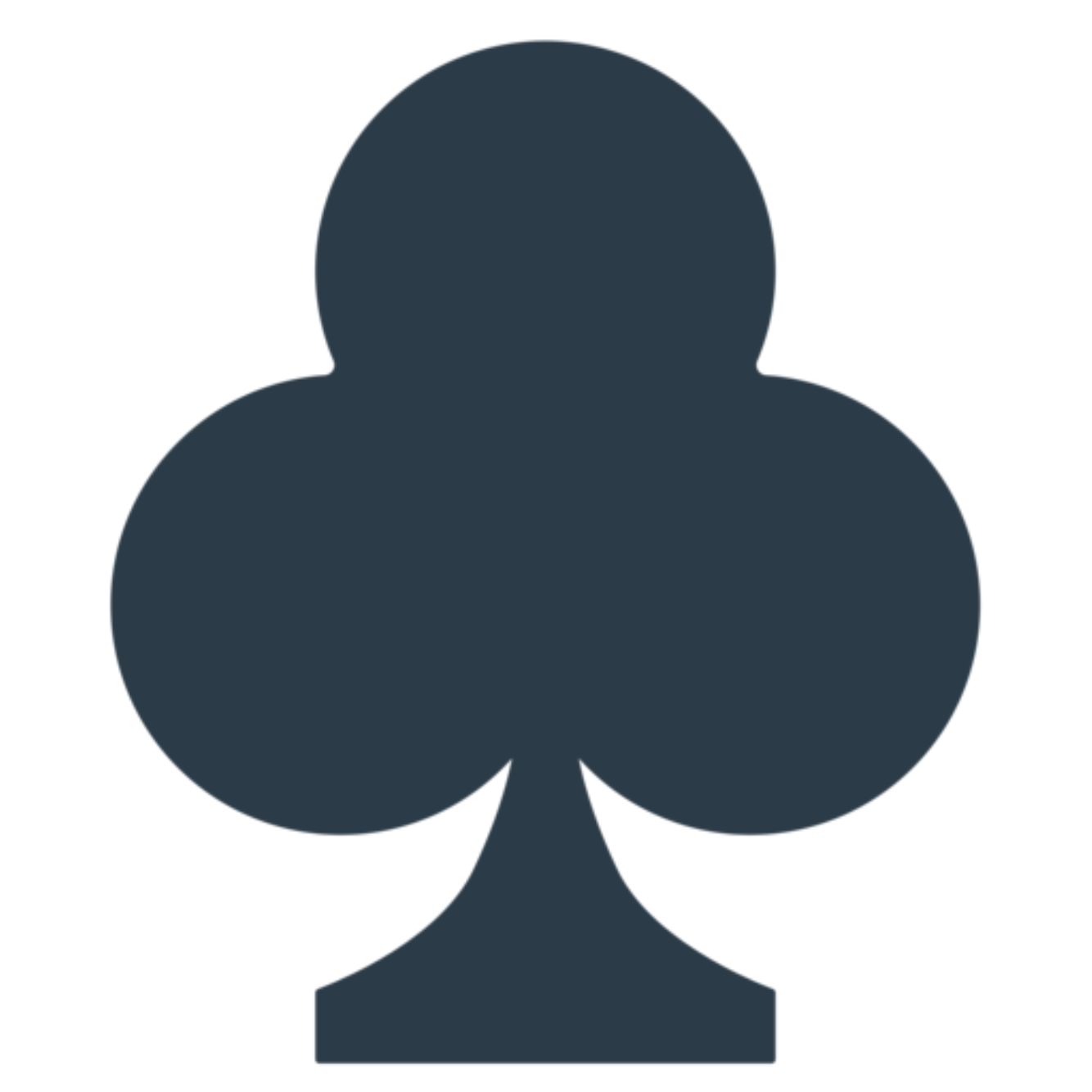}}}

\usepackage{tikz}
\usepackage{soul}

\DeclareRobustCommand{\hlg}[1]{{\sethlcolor{lightgray}\hl{#1}}}
\DeclareRobustCommand{\hlp}[1]{{\sethlcolor{magenta}\hl{#1}}}
\definecolor{cyellow}{HTML}{FFC107}
\DeclareRobustCommand{\hly}[1]{{\sethlcolor{cyellow}\hl{#1}}}

\usepackage[T1]{fontenc}

\usepackage[utf8]{inputenc}

\usepackage{microtype}

\title{Bidimensional Leaderboards:\\ Generate and Evaluate Language Hand in Hand}

\author{First Author \\
  Affiliation / Address line 1 \\
  Affiliation / Address line 2 \\
  Affiliation / Address line 3 \\
  \texttt{email@domain} \\\And
  Second Author \\
  Affiliation / Address line 1 \\
  Affiliation / Address line 2 \\
  Affiliation / Address line 3 \\
  \texttt{email@domain} \\}
  \author{
    Jungo Kasai\handshake\thanks{\ \ Work was done during an internship at AI2.}  
\quad
\textbf{Keisuke Sakaguchi}\heart
\quad 
\textbf{Ronan Le Bras}\heart
\quad
\textbf{Lavinia Dunagan}\handshake
\\
\textbf{Jacob Morrison}\handshake\diamondemoji
\quad 
\textbf{Alexander R.\ Fabbri}\club \quad \textbf{Yejin Choi}\handshake\heart
\quad
\textbf{Noah A.\ Smith}\handshake\heart\\
    \handshake Paul G.\ Allen School of Computer Science \& Engineering, University of Washington
    \\
\heart Allen Institute for AI \quad \diamondemoji Department of Linguistics, University of Washington \\
\club Salesforce Research
    \\
    {\tt billboard.nlp@gmail.com}
}

\begin{document}
\maketitle

\setlength{\abovedisplayskip}{2pt}
\setlength{\belowdisplayskip}{2pt}
\begin{abstract}
Natural language processing researchers have identified limitations of evaluation methodology for generation tasks, with new questions raised about the validity of automatic metrics and of crowdworker judgments.
Meanwhile, efforts to improve \emph{generation models} tend to depend on simple n-gram overlap metrics (e.g., BLEU, ROUGE).
We argue that new advances on models and metrics should each more directly benefit and inform the other.
We therefore propose a generalization of leaderboards, \textbf{bidimensional leaderboards} (\bilboards), that simultaneously tracks progress in language generation models and metrics for their evaluation.
Unlike conventional \textit{unidimensional} leaderboards that sort submitted systems by predetermined metrics, a \bilboard accepts both generators and evaluation metrics as competing entries.
A \bilboard automatically creates an ensemble metric that selects and linearly combines a few metrics based on a global analysis across generators.
Further, metrics are ranked based on their correlation with human judgments.
We release four \bilboards for machine translation, summarization, and image captioning.\footnote{\url{https://nlp.cs.washington.edu/billboard/}.}
We demonstrate that a linear ensemble of a few diverse metrics sometimes substantially outperforms existing metrics in isolation.
Our mixed-effects model analysis shows that most automatic metrics, especially the reference-based ones, overrate machine over human generation, demonstrating the importance of updating metrics as generation models become stronger (and perhaps more similar to humans) in the future. 
\end{abstract}
\section{Introduction}

\begin{figure}[h]
\centering
    \includegraphics[width=0.49\textwidth]{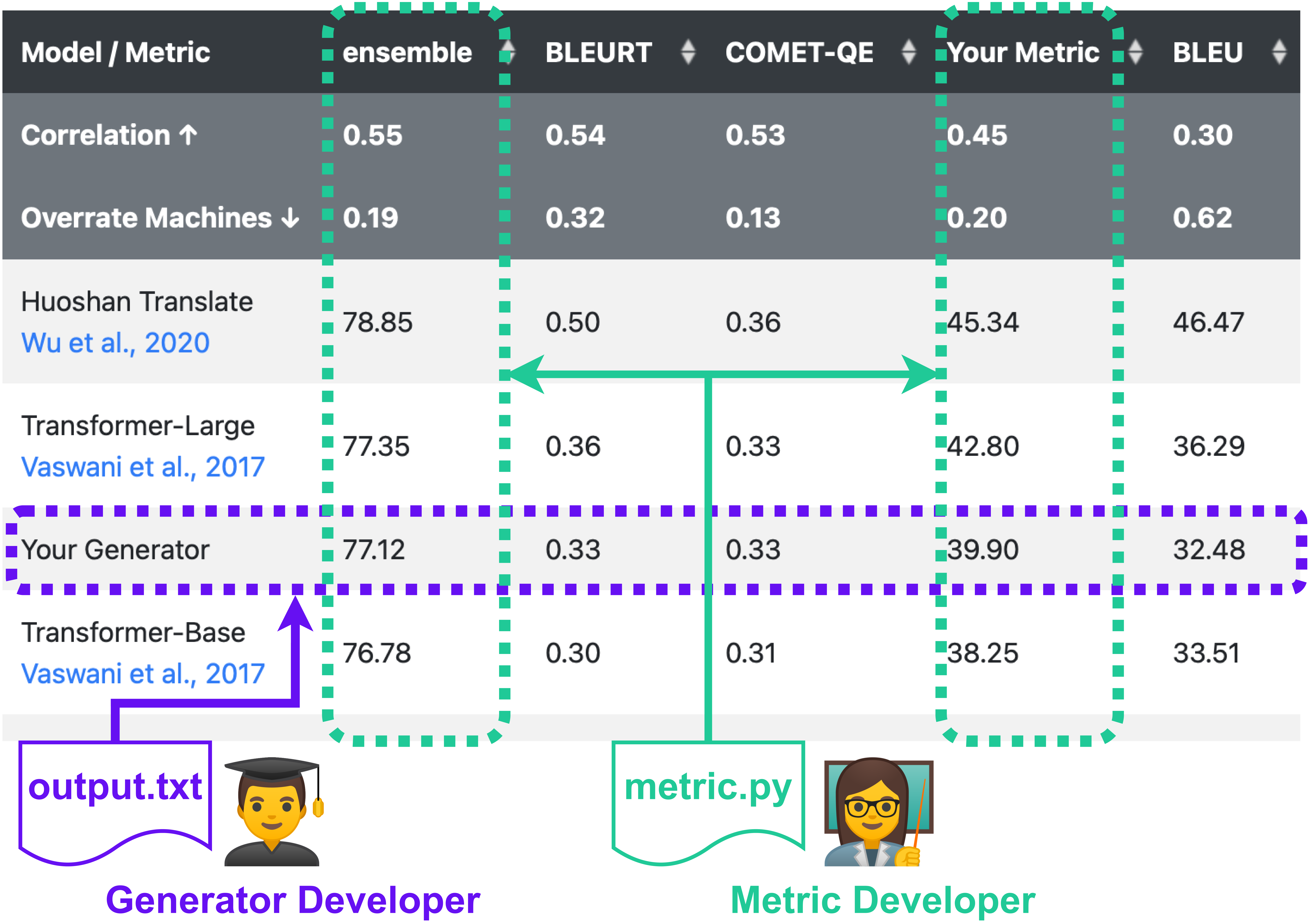}
\caption{Bidimensional leaderboard (\bilboard). When a generator developer submits output text (\texttt{output.txt}), \bilboard computes all metric scores.
When a metric developer submits an executable program (e.g., \texttt{metric.py}), \bilboard computes correlation with the human judgments, updates the ensemble metric (\S\ref{sec:ensemble}), and measures how much the metric overrates machines (\S\ref{sec:overrate}). 
}
\vspace{-0.6cm}
\label{fig:biLB-main}
\end{figure}

Recent modeling advances have led to improved natural language generation in applications such as machine translation and summarization \interalia{ng-etal-2019-facebook, T5, gpt3}.
This progress is typically measured with automatic scores, such as BLEU \cite{Papineni2001BleuAM} and ROUGE \cite{Lin2004ROUGEAP}, executed by modeling researchers themselves.
These metrics allow for fast, inexpensive development cycles.
They were adopted based on reported correlations with human judgments at the time the metrics were introduced, but it has since been established that the correspondence can collapse when models of different types are compared \cite{callison-burch-etal-2006-evaluating} or models become increasingly powerful \cite{ma-etal-2019-results, Edunov2020OnTE}.

Meanwhile, many evaluation metrics that improve correlation with human judgments have been proposed \interalia{clark-etal-2019-sentence, bertscore, sellam2020bleurt, clipscore}, but this progress has yet to be broadly adopted by the community of researchers focused on advancing models.
Indeed, consistent with prior meta-evaluations \cite{marie-etal-2021-scientific}, we found that 68\% of the machine translation papers from NAACL and ACL 2021 evaluated their models solely by BLEU, and only 5\% measured the performance using recent metrics with contextual representations such as COMET \cite{rei-etal-2020-comet}.
Similarly, automatic evaluation in 66\% of the summarization papers was done only in terms of ROUGE.\footnote{We examined all papers whose title contains ``machine translation'' and ``summarization.'' See Appendix \ref{appendix:breakdown} for details.}
We believe this separation between generation modeling and automatic evaluation represents a missed opportunity for each subcommunity to more rapidly benefit from the advances of the other.

We therefore propose an abstraction of conventional leaderboards, \textbf{bidimensional leaderboards} (\bilboards), that simultaneously facilitates progress in natural language generation and its evaluation (Fig.\ \ref{fig:biLB-main}).
A \bilboard accepts two types of submissions related to a given task and dataset: \textbf{generators} and \textbf{metrics}. 
Unlike conventional leaderboards, model ranking is not tied to a predetermined set of metrics; the generators are ranked based on the metric that currently correlates best with human judgments.
Metric submissions are ranked by their correlations to human judgments, and each is stored as an executable program, which will then be used to evaluate future generation submissions.
Our \bilboard includes a sparse regression that selects and linearly combines three existing metrics, revealing  complementary strengths.
All leaderboard scores are readily reproducible, allowing research on generation models and automatic metrics to benefit from each other.

We release four \bilboard interfaces (\url{https://nlp.cs.washington.edu/billboard/}) spanning  three generation tasks: the WMT20 EN-DE and WMT20 ZH-EN machine translation tasks \cite{barrault-etal-2020-findings}, the CNNDM summarization task \cite{cnndaily}, and the MSCOCO image captioning task \cite{mscoco}.

\paragraph{Key Findings}
Using the collective analyses of \bilboards, our main findings are as follows.
\begin{itemize}[noitemsep, leftmargin=*, topsep=1em]
  \setlength\itemsep{1em}
    \item A simple linear combination of a few (diverse) metrics can sometimes improve correlation. This finding quantifies complementary effects of different metrics and encourages metric developers to seek out aspects of generated text quality not yet measured by existing metrics. 
    \item Using linear mixed-effects models, we find that most automatic metrics, especially conventional, reference-based ones such as BLEU and ROUGE, \emph{overrate} machines over humans in all tasks. This result provides further support for the claim that the metrics should be continually evaluated and updated as our generation models become stronger (and perhaps, closer to humans).%
    \item  When only one reference is available per instance, COMET-QE (a strong \textit{referenceless} metric with crosslingual contextual representations; \citealp{rei-etal-2020-comet}) achieves higher correlation with human judgments than all reference-based metrics.
    This raises a concern about the current standard evaluation practice in machine translation and summarization that uses reference-based metrics with a single reference per instance.
    \item Our findings confirm many others who report that recent metrics achieve substantially higher correlation with human judgments than popular metrics like BLEU and ROUGE in \bilboards. We believe these older metrics continue to be used mainly because modeling researchers value consistency and accessibility of evaluation practice over long periods of time.
    \bilboards provide a way to maintain long-term comparability of system output while also drawing better conclusions about system quality, using advances in evaluation.
    All generators continue to be evaluated with new metrics on \bilboards.
\end{itemize}
\section{Bidimensional Leaderboards}\label{sec:bilboard}
We propose \bilboards to simultaneously drive progress in natural language generation and its evaluation, which are often disconnected in current research. 
We first describe the general framework (\S{\ref{sec:bilboard-setup}}) and the automatic analyses they provide (\S{\ref{sec:ensemble}}-\ref{sec:overrate}).
We then discuss our design choices (\S\ref{sec:design-choices}) and the rubric-based, human judgment data necessary to initialize \bilboards (\S\ref{sec:meta-human-evaluation}).

\subsection{\bilboard Framework}
\label{sec:bilboard-setup}
The leaderboard paradigm has driven research on state-of-the-art model performance on many tasks in various fields (e.g., ImageNet, \citealp{imagenet-challenge}; SQuAD, \citealp{rajpurkar-etal-2016-squad}).
As applications and tasks become more diverse, however, the conventional leaderboard paradigm presents a serious challenge: the assumption becomes too strong that predetermined, automatic metrics can reliably score the system performance \textit{over time}.
In particular, scores from automatic metrics often diverge from human judgments in language generation tasks, especially when models become increasingly powerful \cite{ma-etal-2019-results}.

Much recent work proposed new evaluation metrics that improve correlations with human judgments in certain generation tasks \interalia{clark-etal-2019-sentence, bertscore, sellam2020bleurt, clipscore}, but most developers of generation models are not benefiting from them (See Appendix \ref{appendix:breakdown} for our analysis of papers from NAACL/ACL 2021).
From the perspective of generation model developers, it is not clear which of these many metrics in the literature is most reliable in which generation task or dataset, resulting in community-wide overuse of long-standing metrics like BLEU and ROUGE.
Developers of evaluation metrics, on the other hand, are missing the opportunity to apply their metrics to new generation models and compare them with the existing ones.
We propose \bilboards that bridge this gap between generation modeling and evaluation development.

\paragraph{Generators, Metrics, and Scores}
A \bilboard for a language generation task consists of sets of generators and evaluation metrics: $\mathcal{G}= \{G_i \}_{i=1}^I, \mathcal{M}= \{M_j \}_{j=1}^J$.
Each generator $G_i$ takes as input $X_k$ (e.g., source text in machine translation) and generates text: $Y_{i,k} = G_i(X_k)$.
A metric $M_j$ assigns a score to each generated text given the generation input and the corresponding set of references $\mathcal{R}_{k}$: $s_{i,j,k} = M_j(Y_{i,k}, \mathcal{R}_{k}, X_k)$.
The last two arguments to the function are optional; some metrics do not require references (i.e., \textit{referenceless} or \textit{quality estimation} metrics) or the generation input (e.g., BLEU).
We then compute the aggregate score $s_{i,j}$ by averaging $s_{i,j,k}$ over $K$ test examples.

\paragraph{Rankings}
In contrast to standard leaderboards, \bilboards have a dynamic set of evaluation metrics, and generators are not ranked by a predefined metric.
We first rank the metrics by measuring their correlations to human judgments as commonly done in the generation evaluation literature \cite{bertscore, sellam2020bleurt}.
Let $h_{i,k}$ be a human score for $Y_{i,k}$ (i.e., output from generator $G_i$ on input $X_k$).
We compute the instance-level Pearson correlation for every metric $M_j$ between $h_{i,k}$ and $s_{i,j,k}$ ($M_j$ score for $Y_{i,k}$).
All metrics are ranked by their correlations. 
We then use the top metric $M_{j^{*}}$ to rank the generators in the descending order of $s_{i,j^{*}}$.
We defer our discussions on alternative design choices (\S\ref{sec:design-choices}) and human evaluations (\S\ref{sec:meta-human-evaluation}).
We note, however, that the overall framework of \bilboards still holds regardless of these decisions.

\subsection{Ensemble of Metrics}
\label{sec:ensemble}
So far, we have assumed that metrics are used individually in isolation, but \bilboards provide a unique opportunity to examine metrics collectively. 
Different metrics can capture different aspects of generation quality; even if a metric is not sufficiently informative in isolation, it might reflect an important aspect of text quality that the existing metrics overlook.
Here we consider a straightforward and interpretable ensemble of metrics using a regression model with $\ell_1$ regularization \cite{Tibshirani94regressionshrinkage}.
Let the ensemble's score be
\begin{align*}
\hat{h}_{i,k} = \sum_{j=1}^J w_j \cdot s_{i,j,k},
    \end{align*}
    where $w_j$ is a scalar coefficient associated with the $j$th metric and the intercept term is suppressed.
We optimize the vector of coefficients $\mathbf{w}$ with the pairs of output text and a human score $\{Y_{i,k}, h_{i,k}\}_{k=1}^{K}$ from the test data:
\begin{align*}
\mathbf{w^{*}} = \argmin_{\mathbf{w}} \sum_{k=1}^K \left( h_{i,k} - \hat{h}_{i,k}   \right)^2 + \lambda \lVert \mathbf{w} \rVert_{1}
\end{align*}
The $\ell_1$ regularization produces sparse coefficients and improves interpretability by removing highly correlated metrics.
Moreover, it avoids the need for practitioners to run many metrics to obtain an ensemble score when used outside our \bilboards.
Our goal for the ensemble is to provide a useful signal to the research community, rather than to achieve the best possible correlation with human judges at a given time; we tune $\lambda$ to get three non-zero coefficients.
Every metric is standardized by its mean and standard deviation on the test data.

Similar to the individual metrics, we rank this ensemble metric by its correlation to the human judgments.
To make fair comparisons, we simulate situations where the ensemble is applied to a newly submitted generator that has no human evaluations.
Specifically, we perform cross validation that holds out the human judgments for each generator $G_i$ and runs regression on the rest; we then apply these $I$ regression models to the corresponding held-out data and calculate the overall correlation.
We will see that the ensemble metric outperforms all individual metrics in some cases, suggesting that different metrics can capture different aspects.

\paragraph{Reproduciblity}
The ensemble metric is updated every time a new metric is submitted (Fig.\ \ref{fig:biLB-main}). 
For reproducibility, we keep track of every past ensemble metric with a signature that indicates its coefficients, $\lambda$, and input metrics in the backend.
Similar to \textsc{SacreBLEU} \cite{post-2018-call}, model developers can report the signature for easy replication of their scores from the ensemble metric.\footnote{E.g., ensemble.wmt20-zh-en+refs.AB+version.1.}
Further, all generation outputs are saved on the leaderboards, so model developers can download outputs from all past models and compare in any way.

\subsection{Mixed-Effects Model Analysis}
\label{sec:overrate}
Recent work \cite{kasai2021thumb} observed that automatic metrics tend to \textit{overrate} machine-generated text over human one on the MSCOCO image captioning task \cite{mscoco-captioning}.
This problem is particularly severe in conventional metrics that are based on n-gram overlap such as BLEU and CIDEr \cite{cider2015}.
This raises a significant concern about the continuous use of these conventional metrics in generation tasks as models become increasingly powerful (and more similar to humans); those metrics unintentionally discourage researchers from developing human-like, strong generation models.
To quantify this undesirable property, we propose a linear mixed-effects model that compares the two groups of machine- and human-generated text.
The underlying model assumes that $s_{i,j,k}$, the score from metric $M_j$ for generator $G_{i}$ and test example $k$, can be expressed as (the intercept term is suppressed for brevity):
\vspace{0.1cm}
\begin{align*}
s_{i,j,k}\! =\! \beta_{0}^j \mathbbm{1}  \{ G_i\:\text{is machine}\} \! + \! \beta_1^j h_{i,k} \! + \! \gamma_k \! + \! \epsilon_{i,j,k}
\end{align*}
\vspace{-0.3cm}

\noindent where $\gamma_k$ is the random effect for example $k$, and $\epsilon_{i,j,k}$ is Gaussian noise.
Intuitively, $\beta_0^j$ measures how much metric $M_j$ \textit{overrates} machine generation over human one, compared against the human judgment $h_{i,k}$.
$\beta_0^j=0$ means being neutral, and indeed we will find that $\beta_0^j$ is significantly positive in most cases (\S\ref{sec:results}).
We standardize all metric scores over the test samples to compare the size of $\beta_0^j$.
We apply the \emph{lme4} package \cite{lme4-2015}.
\begin{figure*}[h]
\centering
    \includegraphics[width=0.99\textwidth]{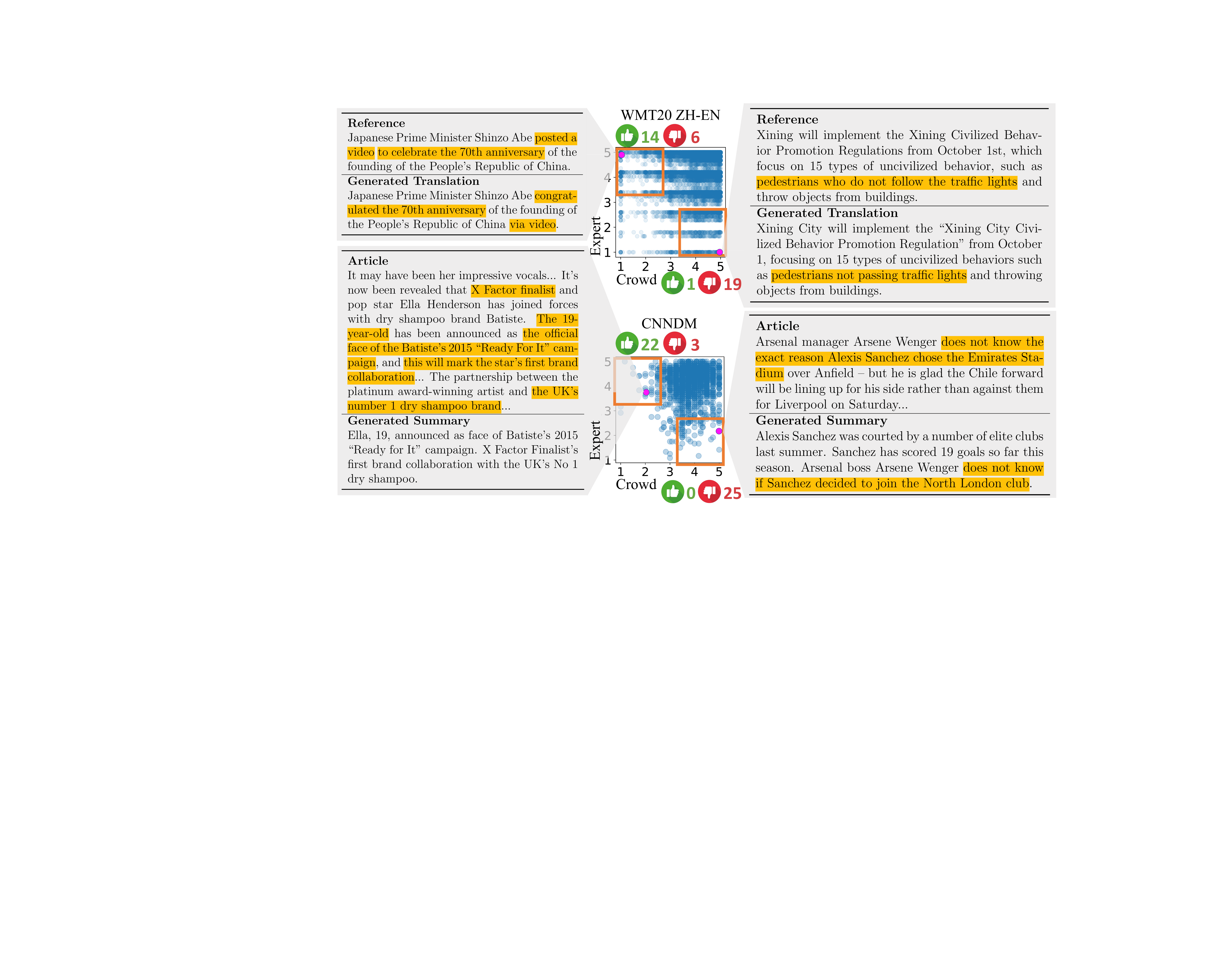}
\caption{
Comparisons and meta-evaluations of crowdworker and rubric-based, expert evaluations for WMT20 ZH-EN and CNNDM summarization.
Every dot represents one test instance that is evaluated by the same numbers of experts and crowdworkers (one for WMT20 ZH-EN and three for CNNDM) for fair comparisons.
We randomly sampled instances with diverging evaluations in two areas \orangesquare~and conducted binary meta-evaluations (good \thumbup~or bad quality \thumbdown). \textbf{Meta-evaluations agree more with the expert evaluations}: \thumbup\ >\ \thumbdown~in the upper left squares and \thumbdown\ >\ \thumbup~in the lower right squares. We suspect that the \hly{highlighted} text might have caused the disagreement.}
\label{fig:expert-crowd}
\end{figure*}

\subsection{Design Choices and Discussion}
\label{sec:design-choices}
In our current setup, we make several design choices for metrics and their rankings:
\vspace{0.2cm}
\begin{itemize}[noitemsep, leftmargin=*, topsep=0pt]
    \item \textbf{M.1} Metrics are expected to positively correlate with the generation output quality.
    \item \textbf{M.2} By default, metrics are ranked based on their instance-level Pearson correlations with human judgments.
    We also compute and present their system-level Kendall rank correlations.
    \item \textbf{M.3} When available, reference-based metrics use multiple references per instance.
\end{itemize}
M.1 implies that we need to take the negative of metric scores that are intended to negatively correlate (e.g., TER, \citealp{TER2006}).
This normalization is also done in WMT metric competitions \interalia{callison-burch-etal-2007-meta, callisonburch-EtAl:2008:WMT}.

While instance-level correlations are commonly used to evaluate and compare automatic metrics for various language generation tasks \interalia{sellam2020bleurt, fabbri2021summeval, clipscore}, there are several alternatives to M.2.
For example, Pearson, Spearman's rank, or Kendall rank correlations can be used on a system (i.e., generator) level \cite{callison-burch-etal-2007-meta, machacek-bojar-2014-results, mathur-etal-2020-results}.
However, such system-level correlations would substantially reduce data points to compare automatic scores, resulting in many ties in the ranking.
Spearman's and Kendall rank correlations become brittle when multiple generators are similar in overall output quality;
penalizing a metric for swapping two similar generators is misleading \cite{machacek-bojar-2014-results}.
Moreover, if a metric can perform well on an instance level, it can be used to augment human judgments by, for example, flagging likely wrong ratings \cite{mathur-etal-2020-results}.
Thus, we encourage researchers to develop metrics that correlate well with human judgments on an instance level.
Prior work also points out other problems in ranking metrics like \textit{outlier effects} where outlier systems have a disproportionately large effect on the overall correlation \cite{tangledup,mathur-etal-2020-results}.
We therefore assume M.2 in the current version of \bilboards, but this can be modified in a future version.

M.3 is supported by our experimental results in \S\ref{sec:results} that multiple references substantially improve reference-based metrics, and a single reference is often insufficient to outperform strong referenceless metrics.
Some metrics have specifications for multiple references (e.g., BLEU, CIDEr). 
In the other cases, we evaluate outputs against every reference and take the maximum score, following prior work on image captioning evaluation \cite{bertscore, clipscore}.\footnote{Intuitively, the maximum score measures the distance to the closest out of equally valid generations.}

\subsection{Human Evaluation}
\label{sec:meta-human-evaluation}
Human evaluations are required to initialize \bilboards; they are used to rank metrics, train the metric ensembling model, and assess how much each metric overrates machines.
Recent work, however, points out problems when evaluations are done by crowdworkers even when extensive quality controls are performed \cite{gillick-liu-2010-non, toral-etal-2018-attaining, freitag2021experts, clark-etal-2021-thats}.
\citet{freitag2021experts} show that rubric-based machine translation evaluations by professional translators led to substantially different generator rankings from the crowdsource evaluations in WMT 2020 \cite{barrault-etal-2020-findings}, where WMT participants or Amazon Mechanical Turkers directly assess each translation's adequacy by a single score (\textit{direct assessment}).
These crowdworker evaluations depend highly on individual annotators’ discretion and understanding of the annotation scheme \cite{freitag2021experts, clark-etal-2021-thats}, making it difficult to decompose, interpret,
and validate \cite{kasai2021thumb}.
Moreover, these direct assessment scores make it difficult to interpret evaluation results for downstream applications where some aspects are particularly important (e.g., accessibility for people with visual impairments in image captioning, \citealp{GleasonCHI20}; gender bias in machine translation, \citealp{stanovsky-etal-2019-evaluating}).

Motivated by this line of work, we perform meta-evaluations to compare crowdsourced and rubric-based expert evaluations.
Fig.\ \ref{fig:expert-crowd} plots overall scores for test examples from WMT20 ZH-EN \cite{barrault-etal-2020-findings, freitag2021experts} and CNNDM summarization \cite{fabbri2021summeval}.
Each instance is evaluated by averaging the same number of crowdworkers and expert scores for fair comparisons.
We see that substantially many instances fall into disagreement: crowdworkers give much higher scores than experts (lower right square) or the reverse (upper left square).
We sample and shuffle 20/25 examples from either type and ask a meta-evaluator to make a binary decision (good \thumbup~or bad quality~\thumbdown).\footnote{The meta-evaluations were done by a bilingual speaker (WMT20 ZH-EN) and the first author of this paper (CNNDM).}
Meta-evaluations agree more with the expert evaluations (e.g., 22 and 0 \thumbup~in the upper left and lower right squares for CNNDM, respectively).
In the examples on the left, crowdworkers fail to properly assess a valid translation with different structure than the reference (\textit{posted a video to celebrate} vs.\ \textit{congratulated via video}) or a summary that combines information from different parts of the article. The examples on the right illustrate that crowdworkers can be fooled by inaccurate yet fluent generations (\textit{does not know the reason} vs.\ \textit{does not know if Sanchez decided}).
Given this result, we decide to initialize our \bilboards with rubric-based expert evaluations for all generation tasks.
We still encourage future work to explore ways to improve crowdsourced evaluations for scalability.
\section{Experiments}

\label{sec:experiments}
Having established the framework, we set up \bilboards for three natural language generation tasks: machine translation, summarization, and image captioning.
To maximize the performance of reference-based metrics, we use as many references as possible for each task.
See \S\ref{sec:results} for an analysis on the effect of varying numbers of references.

\subsection{Tasks}
\paragraph{Machine Translation}
We experiment with two language pairs from the WMT 2020 news translation task \cite{barrault-etal-2020-findings}: Chinese$\rightarrow$English (\textbf{WMT20 ZH-EN}) and English$\rightarrow$German (\textbf{WMT20 EN-DE}).
We use outputs from all submitted translation systems.\footnote{\url{https://www.statmt.org/wmt20/translation-task.html}.}
These two language pairs have expert, rubric-based scores (MQM) from \citet{freitag2021experts} for a subset of 10 submitted systems, including the top-performing systems and human translations.
Each output sentence is evaluated by three professional translators.
Following \citet{freitag2021experts}, the three scores are averaged to get an instance-level score.

We use all human translations available as a reference set for reference-based metrics.
Concretely, every test instance in WMT20 ZH-EN has two translations provided by different human translation services: Human-A and Human-B \cite{barrault-etal-2020-findings}.
In addition to Human-A and Human-B, WMT20 EN-DE provides a translation that is created by linguists who are asked to paraphrase Human-A and Human-B as much as possible (Human-P, \citealp{freitag-bleu-paraphrase-references-2020}).
These paraphrased translations are shown to increase correlations with human judgments by mitigating the \textit{translationese effect} and diversifying the reference when the generation quality is measured by reference-based metrics \cite{freitag-bleu-paraphrase-references-2020}.

\begin{table*}[h!]
\addtolength{\tabcolsep}{-2.0pt}  
\centering
\small
\renewcommand{\arraystretch}{1.3}
\begin{tabular}{@{} lccc|cc|cc @{}}
\toprule
&&&& \multicolumn{2}{c|}{\textbf{Single Metrics}}& \multicolumn{2}{c}{\textbf{Ensemble of Metrics}}\\
Dataset  & $|\mathcal{G} |$ & $|\mathcal{M} |$ & Top Gen.& Top Metric & Corr. & Linear Combination & Corr.\\
 \hline
WMT20 ZH-EN   & 19 &15  & Huoshan & COMET &0.55 & $1.72\cdot$\hlp{COMET-QE}+$1.48\cdot$\hlg{COMET}+$1.21\cdot$\hlg{BLEURT} & \textbf{0.61} \\
WMT20 EN-DE  &17 & 11 & Tohoku & COMET &0.49& $1.19\cdot$\hlg{COMET}+$0.36\cdot$\hlp{COMET-QE}+$0.02\cdot$\hlg{Prism-ref }& \textbf{0.51}\\
CNNDM &  26 & 15 & Lead-3 & COMET & \textbf{0.41} & $2.85\cdot$\hlg{COMET}+$0.26\cdot$\hlp{COMET-QE}+$0.01\cdot$\hlg{BERTScore}& 0.29\\
MSCOCO  &  4 & 15   & VinVL-large & RefCLIP-S & \textbf{0.45} &$2.08\cdot$\hlg{RefCLIP-S}+$1.51\cdot$\hlg{RefOnlyC}+$0.82\cdot$\hlg{CIDEr} & \textbf{0.45} \\
\bottomrule
\end{tabular}
\caption{Summary of \bilboards as of Jan.\ 10, 2022. Huoshan: \citet{wu-etal-2020-volctrans}; Tohoku: \citet{kiyono-etal-2020-tohoku}; VinVL-large: \citet{zhang2021vinvl}; COMET, COMET-QE: \citet{rei-etal-2020-comet}; BLEURT: \citet{sellam2020bleurt}; Prism-ref: \citet{thompson-post-2020-automatic}; BERTScore: \citet{bertscore}; RefCLIP-S: \citet{clipscore}; RefOnlyC: \citet{kasai2021thumb}.
\hlp{COMET-QE} is a \textit{referenceless} metric. BLEURT is specifically trained to evaluate into-English translations. RefCLIP-S uses image features unlike most metrics for image captioning.
RefOnlyC removes image features from RefCLIP-S and only uses reference text features from CLIP \cite{clip2021}.
}
\label{tab:biLB-Summary}
\end{table*}

Along with all submitted generators in WMT20 ZH-EN and WMT20 EN-DE, we train three transformer baselines with the \texttt{fairseq} library \cite{ott-etal-2019-fairseq} and place them in our \bilboards: \textbf{transformer-base}, \textbf{transformer-large}, and \textbf{transformer-large-ensemble} with similar hyperparameters (e.g., 6-layer encoder and decoder) to the ones trained on the WMT16 EN-DE data in \citet{Vaswani2017AttentionIA}.\footnote{Data and models are available at \url{https://github.com/jungokasai/billboard/tree/master/baselines}.}
These baselines allow researchers to compare their translation models without resource-intensive techniques such as backtranslation \cite{sennrich-etal-2016-improving}, model ensembling, and deep encoders \cite{deepshallow}.
These techniques are all used in top-performing systems of WMT20 \cite{wu-etal-2020-volctrans, kiyono-etal-2020-tohoku} but might be infeasible in many research settings.
See Appendix \ref{appendix:generators} for a list of all hyperparameters for the baselines.

\paragraph{Summarization}
We use the CNN/DailyMail corpus (\textbf{CNNDM}, \citealp{cnndaily, nallapati-etal-2016-abstractive}).
We use the standard train/dev./test split and 24 models from \citet{fabbri2021summeval}.
100 test articles are annotated with 10 summaries written by humans \cite{kryscinski-etal-2019-neural}.
For those 100 articles, rubric-based, expert evaluations for 18 generators, including human-written highlights, are provided by \citet{fabbri2021summeval}.\footnote{Some of the outputs are lowercased and/or tokenized. In these cases, we apply the NLTK detokenizer \cite{nltk} and/or the Stanford CoreNLP truecaser \cite{manning-EtAl:2014:P14-5}. We encourage, however, future model developers to provide clean, untokenized output to improve the
reproducibility and transparency of evaluation results \cite{post-2018-call,kasai2021thumb}.}
Each output summary is evaluated by three experts along four dimensions: \textit{coherence} (collective quality of all summary sentences), \textit{consistency} (factual alignment with the article, penalizing for hallucinations), \textit{fluency} (quality of the individual sentences), and \textit{relevance} (selection of important content).
An instance-level score is computed by averaging scores over all these categories and the three experts.
Note that this aggregation method can be modified, depending on the downstream task of interest \cite{kasai2021thumb}.
All 10 human-written summaries are used as the reference set for reference-based metrics.\footnote{Prior work used a concatenation of author-written highlights as a reference, but here we do not add it to the reference set. This is because these highlights are sometimes noisy (e.g., containing URLs) or lack coherence \cite{fabbri2021summeval}.}

\paragraph{Image Captioning}
We use the \textbf{MSCOCO} dataset \cite{mscoco} that consists of everyday-scene photos sampled from Flickr.
Every image is annotated with five captions written by crowdworkers \cite{mscoco-captioning}.
We apply the standard \textit{Karpathy split} \cite{Karpathy_2015_CVPR}.
For each of 500 test images, rubric-based evaluations (\thumb 1.0) are available for five systems, including one caption from a crowdworker \cite{kasai2021thumb}.
Similar to machine translation and summarization, we use all five crowdworker captions as a reference set for reference-based metrics.

\subsection{Mixed-Effects Models}
Our mixed-effects model analyzes how much every automatic metric overrates machines over humans (\S\ref{sec:overrate}).
This means that we need to free up one human generation per instance to measure its scores in the reference-based metrics.
For machine translation, we score Human-B using the reference set of Human-A (WMT20 ZH-EN) or Human-A and Human-P (WMT20 EN-DE).
For CNNDM, we use concatenated highlights as human-generated summaries and use the 10 human-written summaries from \citet{kryscinski-etal-2019-neural} as the reference.
We follow \citet{kasai2021thumb} for MSCOCO and score their randomly-selected \textit{Human} caption using the other four as the reference.
As the distinction between the \textit{reference} and \textit{human generation} (e.g., Human-A vs.\ Human B on WMT20 ZH-EN) is arbitrary, we found that swapping the roles would still lead to similar results (See Appendix \ref{appendix:mixed-effect}).

\section{Results and Analysis}
\label{sec:results}
Here we discuss the current results and make several key observations about the state of language generation evaluation.
Table \ref{tab:biLB-Summary} summarizes the four \bilboards.
It is particularly noteworthy that COMET, a metric designed for machine translation, achieves the best correlation on the CNNDM summarization task as well.
COMET evaluates the similarity between the crosslingual representations from XLM-RoBERTa \cite{conneau-etal-2020-unsupervised} for input text and its translation candidate.
But these crosslingual representations can, of course, be used \textit{monolingually} for English summarization.
This illustrates an additional benefit of \bilboards that centralize different generation tasks and find surprising task transferability of learning-based metrics.
See Appendices \ref{appendix:generators} and \ref{appendix:metrics} for lists of all participating generators and metrics.

\paragraph{Ensemble Metric}
The rightmost section of Table~\ref{tab:biLB-Summary} shows the chosen metrics and their coefficients in the ensemble (\S\ref{sec:ensemble}).
On the machine translation tasks, the ensemble metric outperforms the top individual metric.\footnote{We found a major reason for the anomaly in CNNDM; an outlier generator that does not use the standard CNNDM training data (the GPT-2 zero-shot model; \citealp{gpt2-zero-shot-2019}) has a disproportionately large effect on the regression models. The ensemble metric outperformed the top individual metric of COMET when the zero-shot model was removed.
} 
In particular, we see a substantial gain of $0.06$ points in WMT20 ZH-EN.
The \textit{referenceless} metric of COMET-QE is selected both for WMT20 ZH-EN and WMT20 EN-DE, suggesting complementary effects of diverse metrics.
To further test this hypothesis, we perform ablations that drop one out of the three metrics at a time (Table \ref{tab:ensemble-ablations}).
We see that only dropping COMET-QE would result in a decrease in the correlation score. This implies that the referenceless metric provides important information that the others do not.
\begin{table}[h]
\centering
\small
\addtolength{\tabcolsep}{-4.0pt}  
\renewcommand{\arraystretch}{1.3}
\begin{tabular}{@{} ccccc @{}}
\toprule
 Removed Metric &-- & COMET & \hlp{COMET-QE} & BLEURT \\ 
 \midrule 
 Correlation & 0.61 & 0.61 & 0.57 & 0.61\\
 \bottomrule
\end{tabular}
\caption{Ensemble ablation studies on WMT20 ZH-EN. Only removing COMET-QE leads to a correlation drop. See Appendix \ref{appendix:ensemble-ablations} for the other datasets.}
\label{tab:ensemble-ablations}
\end{table}

\paragraph{Mixed-Effects Models}
Seen in Table \ref{tab:random-effects} are the results from our analysis that measures how much metrics \textit{overrate} machines over humans (\S\ref{sec:overrate}).
We see that the fixed-effect coefficient $\beta_0$ is significantly positive in most cases.
Referenceless metrics tend to have smaller coefficients.
This can be due to the more diverse nature of human text than machine-generated text; reference-based metrics give a low score to human text that differs from the references even if it is of high quality.
The conventional n-gram overlap-based metrics (BLEU, ROUGE, and CIDEr) have particularly large coefficients.
These results suggest that the evaluation practice should be regularly updated as our generation models become stronger (and perhaps, more similar to human generation) in the future.
Note that unlike the other tasks, ``human-generated text'' for CNNDM summarization is an automatic concatenation of author highlights, which contains substantial noise \cite{fabbri2021summeval}. This might explain the neutral and negative coefficients.

\begin{table}[h]
\vspace{0.3cm}
\centering
\small
\addtolength{\tabcolsep}{-5.3pt}  
\renewcommand{\arraystretch}{1.3}
\begin{tabular}{@{} cccccc @{}}
\toprule
\multirow{2}{*}{\textbf{ZH-EN}}  & COMET & \hlp{COMET-QE} & BLEURT & BLEU \\
 & \textred{$0.27_{\pm0.02}$} & \textred{$0.13_{\pm0.01}$} & \textred{$0.32_{\pm0.02}$} & \textred{$0.62_{\pm0.02}$} \\
 \midrule
\multirow{2}{*}{\textbf{EN-DE}}  & COMET & \hlp{COMET-QE} & Prism-ref & BLEU \\
 & \textred{$0.08_{\pm0.03}$} & \textblue{$-0.17_{\pm0.02}$} & \textred{$0.44_{\pm0.02}$} & \textred{$0.33_{\pm0.03}$} \\
 \midrule
\multirow{2}{*}{\textbf{CNNDM}}  & COMET & \hlp{COMET-QE} & BERTScore & ROUGE-L \\
 & \textblue{$-0.17_{\pm0.12}$} & $0.02_{\pm0.11}$ & $-0.04_{\pm 0.12}$ & \textred{$0.33_{\pm0.13}$} \\
 \midrule
\multirow{2}{*}{\textbf{COCO}}  & RefCLIP-S & RefOnlyC & CIDEr& \hlp{CLIP-S} \\
 & \textred{$0.09_{\pm0.06}$} & \textred{$0.24_{\pm0.06}$} & \textred{$0.43_{\pm 0.06}$} & ${-0.04_{\pm0.05}}$ \\
\bottomrule
\end{tabular}
\caption{$\beta_{0}$ fixed-effect coefficients from the linear mixed-effects models, quantifying how much automatic metrics \textbf{overrate} machines over humans, relative to human raters. $\beta_0\!=\!0$ is neutral, and statistical significance is indicated by \textred{red} (positive) or \textblue{blue} text (negative). The subscripts indicate 90\% confidence intervals. Three metrics that correlate best with the human judgments are shown as well as one popular metric. \hlp{COMET-QE} and \hlp{CLIP-S} are \textit{referenceless}. See \S\ref{appendix:mixed-effect} for the other metrics.}
\vspace{-0.0cm}
\label{tab:random-effects}
\end{table}

\begin{figure}[h]
\centering
    \includegraphics[width=0.499\textwidth]{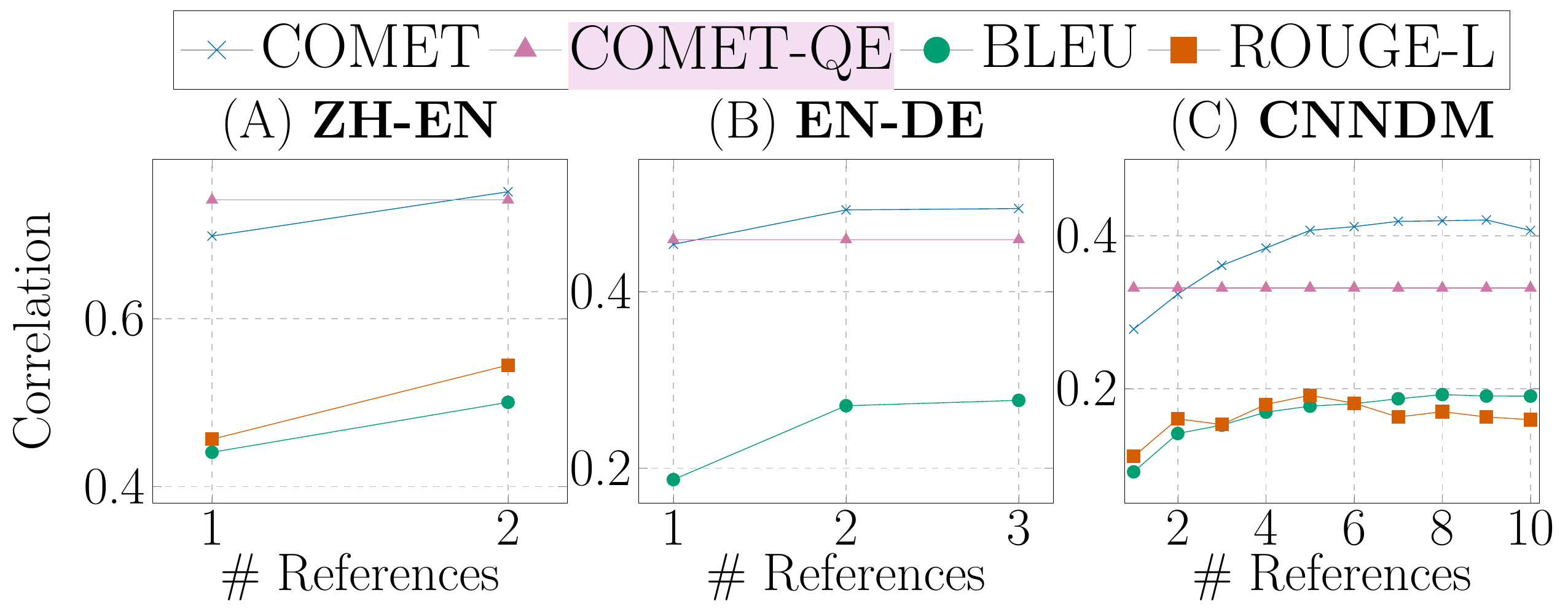}
\caption{Correlations with varying numbers of references. In all cases, one reference is not sufficient to outperform the referenceless \hlp{COMET-QE} metric. The default ROUGE assumes English input.}
\label{fig:nb-references}
\end{figure}

\paragraph{Effects of the Number of References}
Fig.\ \ref{fig:nb-references} plots correlations over varying numbers of references.
COMET was the top-performing \textit{reference-based} metric regardless of the number of references, but we observe that it underperforms the refererenceless metric when only one reference is given.
Model performance in machine translation and summarization is commonly measured by applying reference-based metrics against one reference per instance in the research community.
Our finding thus raises a further concern about the current evaluation practice.
Finally, we see that popular choices of BLEU and ROUGE metrics have much lower correlations than the recent metrics over various numbers of references, in line with the recent studies \interalia{tangledup}.

\section{Related and Future Work}
\paragraph{Related Benchmarks}
WMT organizes the metric competition track in parallel with the translation task every year \interalia{mathur-etal-2020-results, barrault-etal-2020-findings}.
Participants submit automatic scores for the translation outputs from the parallel translation task.
Unfortunately, most of these new metrics are not used by subsequent machine translation work, perhaps because they are tested solely against the concurrent translation submissions and it is up to model developers to execute or even implement new metrics.
The GEM workshop \cite{gem2021} conducts extensive analysis of models and evaluation methods over a wide set of generation tasks.
\bilboards ease the burden through \textit{standard} leaderboard experience where generator developers only need to upload generation outputs for the test split.
\bilboards also offer automatic ensembling of metrics and quantify the diversity that a new metric adds.
The human-in-the-loop \genie leaderboard \cite{genie} centralizes crowdsourced evaluations for generation tasks.
The current \bilboard setup is based on rubric-based, expert evaluation data from previous work, but future work can explore ways to improve crowdsourced evaluations and use them to update \bilboards.

\paragraph{From Bidimensional to Multidimensional}
\bilboards lend themselves to a natural extension: \textit{multidimensional leaderboards}.
In particular, generation models have more aspects than generation quality, such as training and inference efficiency, sample efficiency, and robustness.
These aspects are often ignored in the current leaderboard paradigm but are important to better serving practitioners' needs \cite{greenai, ethayarajh-jurafsky-2020-utility,MishraA21}.
There are ongoing modeling and benchmarking efforts especially for efficient machine translation \interalia{wngt2020, peng2021rfa,kasai2021t2r}.
We leave this extension to future work and specifically target the gap between generation modeling and evaluation.

\section{Conclusion}\label{sec:conclusion}
We introduced \bilboards, a simple yet powerful generalization of leaderboards that bridges the gap between generation modeling and evaluation research.
We established and released four \bilboards on machine translation, summarization, and image captioning tasks.
We demonstrated that their built-in analysis of metric ensembling and mixed-effects modeling revealed key insights into the current state of natural language generation and its evaluation methods.
\bilboards allow for a standard leaderboard experience both on the modeling and evaluation sides.
We invite submissions from researchers through our website.

\section*{Acknowledgements}
We thank Kyunghyun Cho, Elizabeth Clark, Jesse Dodge, Saadia Gabriel, Michal Guerquin, Daniel Khashabi, Swaroop Mishra, Jamie Morgenstern, Phoebe Mulcaire, Hao Peng, Sofia Serrano, the ARK group at UW, the Mosaic team at the Allen Institute for AI, and the anonymous reviewers for their helpful feedback on this work.
We also thank Xinyan Yu for WMT Chinese-English meta-evaluations and Muyun Yang and Shujian Huang for sharing with us the CWMT training data. 
This work was supported in part by the DARPA MCS program through NIWC Pacific (N66001-19-2-4031) and Google Cloud Compute. 

\bibliography{acl}

\begin{thebibliography}{110}
\expandafter\ifx\csname natexlab\endcsname\relax\def\natexlab#1{#1}\fi

\bibitem[{Anderson et~al.(2016)Anderson, Fernando, Johnson, and
  Gould}]{SPICE16}
Peter Anderson, Basura Fernando, Mark Johnson, and Stephen Gould. 2016.
\newblock \href {https://arxiv.org/abs/1607.08822} {{SPICE:} semantic
  propositional image caption evaluation}.
\newblock In \emph{Proc.\ of ECCV}.

\bibitem[{Anderson et~al.(2018)Anderson, He, Buehler, Teney, Johnson, Gould,
  and Zhang}]{Anderson2017up-down}
Peter Anderson, Xiaodong He, Chris Buehler, Damien Teney, Mark Johnson, Stephen
  Gould, and Lei Zhang. 2018.
\newblock \href {https://arxiv.org/abs/1707.07998} {Bottom-up and top-down
  attention for image captioning and visual question answering}.
\newblock In \emph{Proc.\ of CVPR}.

\bibitem[{Banerjee and Lavie(2005)}]{banerjee-lavie-2005-meteor}
Satanjeev Banerjee and Alon Lavie. 2005.
\newblock \href {https://aclanthology.org/W05-0909} {{METEOR}: An automatic
  metric for {MT} evaluation with improved correlation with human judgments}.
\newblock In \emph{Proc.\ of the {ACL} Workshop on Intrinsic and Extrinsic
  Evaluation Measures for Machine Translation and/or Summarization}.

\bibitem[{Barrault et~al.(2020)Barrault, Biesialska, Bojar, Costa-juss{\`a},
  Federmann, Graham, Grundkiewicz, Haddow, Huck, Joanis, Kocmi, Koehn, Lo,
  Ljube{\v{s}}i{\'c}, Monz, Morishita, Nagata, Nakazawa, Pal, Post, and
  Zampieri}]{barrault-etal-2020-findings}
Lo{\"\i}c Barrault, Magdalena Biesialska, Ond{\v{r}}ej Bojar, Marta~R.
  Costa-juss{\`a}, Christian Federmann, Yvette Graham, Roman Grundkiewicz,
  Barry Haddow, Matthias Huck, Eric Joanis, Tom Kocmi, Philipp Koehn, Chi-kiu
  Lo, Nikola Ljube{\v{s}}i{\'c}, Christof Monz, Makoto Morishita, Masaaki
  Nagata, Toshiaki Nakazawa, Santanu Pal, Matt Post, and Marcos Zampieri. 2020.
\newblock \href {https://aclanthology.org/2020.wmt-1.1} {Findings of the 2020
  conference on machine translation ({WMT}20)}.
\newblock In \emph{Proc.\ of WMT}.

\bibitem[{Bates et~al.(2015)Bates, M{\"a}chler, Bolker, and Walker}]{lme4-2015}
Douglas Bates, Martin M{\"a}chler, Ben Bolker, and Steve Walker. 2015.
\newblock \href {https://www.jstatsoft.org/index.php/jss/article/view/v067i01}
  {Fitting linear mixed-effects models using lme4}.
\newblock \emph{Journal of Statistical Software}.

\bibitem[{Bawden et~al.(2020)Bawden, Di~Nunzio, Grozea, Jauregi~Unanue,
  Jimeno~Yepes, Mah, Martinez, N{\'e}v{\'e}ol, Neves, Oronoz, Perez-de
  Vi{\~n}aspre, Piccardi, Roller, Siu, Thomas, Vezzani, Vicente~Navarro,
  Wiemann, and Yeganova}]{bawden-etal-2020-findings}
Rachel Bawden, Giorgio~Maria Di~Nunzio, Cristian Grozea, Inigo Jauregi~Unanue,
  Antonio Jimeno~Yepes, Nancy Mah, David Martinez, Aur{\'e}lie N{\'e}v{\'e}ol,
  Mariana Neves, Maite Oronoz, Olatz Perez-de Vi{\~n}aspre, Massimo Piccardi,
  Roland Roller, Amy Siu, Philippe Thomas, Federica Vezzani, Maika
  Vicente~Navarro, Dina Wiemann, and Lana Yeganova. 2020.
\newblock \href {https://aclanthology.org/2020.wmt-1.76} {Findings of the {WMT}
  2020 biomedical translation shared task: {B}asque, {I}talian and {R}ussian as
  new additional languages}.
\newblock In \emph{Proc.\ of WMT}.

\bibitem[{Bird et~al.(2009)Bird, Klein, and Loper}]{nltk}
Steven Bird, Evan Klein, and Edward Loper. 2009.
\newblock \href {https://doi.org/10.1017/S1351324910000306} {\emph{Natural
  Language Processing with {P}ython}}.
\newblock Cambridge University Press.

\bibitem[{B{\"o}hm et~al.(2019)B{\"o}hm, Gao, Meyer, Shapira, Dagan, and
  Gurevych}]{bohm-etal-2019-better}
Florian B{\"o}hm, Yang Gao, Christian~M. Meyer, Ori Shapira, Ido Dagan, and
  Iryna Gurevych. 2019.
\newblock \href {https://arxiv.org/abs/1909.01214} {Better rewards yield better
  summaries: Learning to summarise without references}.
\newblock In \emph{Proc.\ of EMNLP}.

\bibitem[{Bouscarrat et~al.(2019)Bouscarrat, Bonnefoy, Peel, and
  Pereira}]{bouscarrat-etal-2019-strass}
L{\'e}o Bouscarrat, Antoine Bonnefoy, Thomas Peel, and C{\'e}cile Pereira.
  2019.
\newblock \href {https://arxiv.org/abs/1907.07323} {{STRASS}: A light and
  effective method for extractive summarization based on sentence embeddings}.
\newblock In \emph{Proc.\ of ACL}.

\bibitem[{Brown et~al.(2020)Brown, Mann, Ryder, Subbiah, Kaplan, Dhariwal,
  Neelakantan, Shyam, Sastry, Askell, Agarwal, Herbert-Voss, Krueger, Henighan,
  Child, Ramesh, Ziegler, Wu, Winter, Hesse, Chen, Sigler, Litwin, Gray, Chess,
  Clark, Berner, McCandlish, Radford, Sutskever, and Amodei}]{gpt3}
Tom Brown, Benjamin Mann, Nick Ryder, Melanie Subbiah, Jared~D Kaplan, Prafulla
  Dhariwal, Arvind Neelakantan, Pranav Shyam, Girish Sastry, Amanda Askell,
  Sandhini Agarwal, Ariel Herbert-Voss, Gretchen Krueger, Tom Henighan, Rewon
  Child, Aditya Ramesh, Daniel Ziegler, Jeffrey Wu, Clemens Winter, Chris
  Hesse, Mark Chen, Eric Sigler, Mateusz Litwin, Scott Gray, Benjamin Chess,
  Jack Clark, Christopher Berner, Sam McCandlish, Alec Radford, Ilya Sutskever,
  and Dario Amodei. 2020.
\newblock \href
  {https://proceedings.neurips.cc/paper/2020/file/1457c0d6bfcb4967418bfb8ac142f64a-Paper.pdf}
  {Language models are few-shot learners}.
\newblock In \emph{Proc.\ of NeurIPS}.

\bibitem[{Callison-Burch et~al.(2007)Callison-Burch, Fordyce, Koehn, Monz, and
  Schroeder}]{callison-burch-etal-2007-meta}
Chris Callison-Burch, Cameron Fordyce, Philipp Koehn, Christof Monz, and Josh
  Schroeder. 2007.
\newblock \href {https://aclanthology.org/W07-0718} {(meta-) evaluation of
  machine translation}.
\newblock In \emph{Proc.\ of WMT}.

\bibitem[{Callison-Burch et~al.(2008)Callison-Burch, Fordyce, Koehn, Monz, and
  Schroeder}]{callisonburch-EtAl:2008:WMT}
Chris Callison-Burch, Cameron Fordyce, Philipp Koehn, Christof Monz, and Josh
  Schroeder. 2008.
\newblock \href {http://www.aclweb.org/anthology/W/W08/W08-0309} {Further
  meta-evaluation of machine translation}.
\newblock In \emph{Proc.\ of WMT}.

\bibitem[{Callison-Burch et~al.(2006)Callison-Burch, Osborne, and
  Koehn}]{callison-burch-etal-2006-evaluating}
Chris Callison-Burch, Miles Osborne, and Philipp Koehn. 2006.
\newblock \href {https://www.aclweb.org/anthology/E06-1032} {Re-evaluating the
  role of {B}leu in machine translation research}.
\newblock In \emph{Proc.\ of EACL}.

\bibitem[{Chen et~al.(2020)Chen, Wang, Wei, Shi, Li, Ye, and
  Knight}]{chen-etal-2020-didis}
Tanfang Chen, Weiwei Wang, Wenyang Wei, Xing Shi, Xiangang Li, Jieping Ye, and
  Kevin Knight. 2020.
\newblock \href {https://arxiv.org/abs/2010.08185} {{D}i{D}i{'}s machine
  translation system for {WMT}2020}.
\newblock In \emph{Proc.\ of WMT}.

\bibitem[{Chen et~al.(2015)Chen, Fang, Lin, Vedantam, Gupta, Doll{\'{a}}r, and
  Zitnick}]{mscoco-captioning}
Xinlei Chen, Hao Fang, Tsung{-}Yi Lin, Ramakrishna Vedantam, Saurabh Gupta,
  Piotr Doll{\'{a}}r, and C.~Lawrence Zitnick. 2015.
\newblock \href {http://arxiv.org/abs/1504.00325} {Microsoft {COCO} captions:
  Data collection and evaluation server}.

\bibitem[{Chen and Bansal(2018)}]{chen-bansal-2018-fast}
Yen-Chun Chen and Mohit Bansal. 2018.
\newblock \href {https://arxiv.org/abs/1805.11080} {Fast abstractive
  summarization with reinforce-selected sentence rewriting}.
\newblock In \emph{Proc.\ of ACL}.

\bibitem[{Clark et~al.(2021)Clark, August, Serrano, Haduong, Gururangan, and
  Smith}]{clark-etal-2021-thats}
Elizabeth Clark, Tal August, Sofia Serrano, Nikita Haduong, Suchin Gururangan,
  and Noah~A. Smith. 2021.
\newblock \href {https://arxiv.org/abs/2107.00061} {All that{'}s {`}human{'} is
  not gold: Evaluating human evaluation of generated text}.
\newblock In \emph{Proc.\ of ACL}.

\bibitem[{Clark et~al.(2019)Clark, Celikyilmaz, and
  Smith}]{clark-etal-2019-sentence}
Elizabeth Clark, Asli Celikyilmaz, and Noah~A. Smith. 2019.
\newblock \href {https://aclanthology.org/P19-1264/} {Sentence mover{'}s
  similarity: Automatic evaluation for multi-sentence texts}.
\newblock In \emph{Proc.\ of ACL}.

\bibitem[{Conneau et~al.(2020)Conneau, Khandelwal, Goyal, Chaudhary, Wenzek,
  Guzm{\'a}n, Grave, Ott, Zettlemoyer, and
  Stoyanov}]{conneau-etal-2020-unsupervised}
Alexis Conneau, Kartikay Khandelwal, Naman Goyal, Vishrav Chaudhary, Guillaume
  Wenzek, Francisco Guzm{\'a}n, Edouard Grave, Myle Ott, Luke Zettlemoyer, and
  Veselin Stoyanov. 2020.
\newblock \href {https://arxiv.org/abs/1911.02116} {Unsupervised cross-lingual
  representation learning at scale}.
\newblock In \emph{Proc.\ of ACL}.

\bibitem[{Devlin et~al.(2019)Devlin, Chang, Lee, and
  Toutanova}]{devlins2019bert}
Jacob Devlin, Ming-Wei Chang, Kenton Lee, and Kristina Toutanova. 2019.
\newblock \href {https://arxiv.org/abs/810.04805} {{BERT}: Pre-training of deep
  bidirectional transformers for language understanding}.
\newblock In \emph{Proc. of NAACL}.

\bibitem[{Dong et~al.(2019)Dong, Yang, Wang, Wei, Liu, Wang, Gao, Zhou, and
  Hon}]{UniLM}
Li~Dong, Nan Yang, Wenhui Wang, Furu Wei, Xiaodong Liu, Yu~Wang, Jianfeng Gao,
  Ming Zhou, and Hsiao-Wuen Hon. 2019.
\newblock \href {https://arxiv.org/abs/1905.03197} {Unified language model
  pre-training for natural language understanding and generation}.
\newblock In \emph{Proc.\ of NeurIPS}.

\bibitem[{Dong et~al.(2018)Dong, Shen, Crawford, van Hoof, and
  Cheung}]{dong-etal-2018-banditsum}
Yue Dong, Yikang Shen, Eric Crawford, Herke van Hoof, and Jackie Chi~Kit
  Cheung. 2018.
\newblock \href {https://arxiv.org/abs/1809.09672} {{B}andit{S}um: Extractive
  summarization as a contextual bandit}.
\newblock In \emph{Proc.\ of EMNLP}.

\bibitem[{Edunov et~al.(2020)Edunov, Ott, Ranzato, and Auli}]{Edunov2020OnTE}
Sergey Edunov, Myle Ott, Marc'Aurelio Ranzato, and Michael Auli. 2020.
\newblock \href {https://arxiv.org/abs/1908.05204} {On the evaluation of
  machine translation systems trained with back-translation}.
\newblock In \emph{Proc.\ of ACL}.

\bibitem[{Ethayarajh and Jurafsky(2020)}]{ethayarajh-jurafsky-2020-utility}
Kawin Ethayarajh and Dan Jurafsky. 2020.
\newblock \href {https://arxiv.org/abs/2009.13888} {Utility is in the eye of
  the user: A critique of {NLP} leaderboards}.
\newblock In \emph{Proc.\ of EMNLP}.

\bibitem[{Eyal et~al.(2019)Eyal, Baumel, and Elhadad}]{eyal-etal-2019-question}
Matan Eyal, Tal Baumel, and Michael Elhadad. 2019.
\newblock \href {https://arxiv.org/abs/1906.00318} {Question answering as an
  automatic evaluation metric for news article summarization}.
\newblock In \emph{Proc.\ of NAACL}.

\bibitem[{Fabbri et~al.(2021)Fabbri, Kry{\'s}ci{\'n}ski, McCann, Xiong, Socher,
  and Radev}]{fabbri2021summeval}
Alexander~R Fabbri, Wojciech Kry{\'s}ci{\'n}ski, Bryan McCann, Caiming Xiong,
  Richard Socher, and Dragomir Radev. 2021.
\newblock \href {https://arxiv.org/abs/2007.12626} {{SummEval}: Re-evaluating
  summarization evaluation}.
\newblock \emph{TACL}.

\bibitem[{Freitag et~al.(2021)Freitag, Foster, Grangier, Ratnakar, Tan, and
  Macherey}]{freitag2021experts}
Markus Freitag, George Foster, David Grangier, Viresh Ratnakar, Qijun Tan, and
  Wolfgang Macherey. 2021.
\newblock \href {https://arxiv.org/abs/2104.14478} {Experts, errors, and
  context: A large-scale study of human evaluation for machine translation}.
\newblock \emph{TACL}.

\bibitem[{Freitag et~al.(2020)Freitag, Grangier, and
  Caswell}]{freitag-bleu-paraphrase-references-2020}
Markus Freitag, David Grangier, and Isaac Caswell. 2020.
\newblock \href {https://arxiv.org/abs/2004.06063} {{BLEU} might be guilty but
  references are not innocent}.
\newblock In \emph{Proc.\ of EMNLP}.

\bibitem[{Gehrmann et~al.(2021)Gehrmann, Adewumi, Aggarwal, Ammanamanchi,
  Anuoluwapo, Bosselut, Chandu, Clinciu, Das, Dhole, Du, Durmus, Dusek, Emezue,
  Gangal, Garbacea, Hashimoto, Hou, Jernite, Jhamtani, Ji, Jolly, Kumar,
  Ladhak, Madaan, Maddela, Mahajan, Mahamood, Majumder, Martins,
  McMillan{-}Major, Mille, van Miltenburg, Nadeem, Narayan, Nikolaev,
  Niyongabo, Osei, Parikh, Perez{-}Beltrachini, Rao, Raunak, Rodriguez,
  Santhanam, Sedoc, Sellam, Shaikh, Shimorina, Cabezudo, Strobelt, Subramani,
  Xu, Yang, Yerukola, and Zhou}]{gem2021}
Sebastian Gehrmann, Tosin~P. Adewumi, Karmanya Aggarwal, Pawan~Sasanka
  Ammanamanchi, Aremu Anuoluwapo, Antoine Bosselut, Khyathi~Raghavi Chandu,
  Miruna{-}Adriana Clinciu, Dipanjan Das, Kaustubh~D. Dhole, Wanyu Du, Esin
  Durmus, Ondrej Dusek, Chris Emezue, Varun Gangal, Cristina Garbacea,
  Tatsunori Hashimoto, Yufang Hou, Yacine Jernite, Harsh Jhamtani, Yangfeng Ji,
  Shailza Jolly, Dhruv Kumar, Faisal Ladhak, Aman Madaan, Mounica Maddela,
  Khyati Mahajan, Saad Mahamood, Bodhisattwa~Prasad Majumder, Pedro~Henrique
  Martins, Angelina McMillan{-}Major, Simon Mille, Emiel van Miltenburg, Moin
  Nadeem, Shashi Narayan, Vitaly Nikolaev, Rubungo~Andre Niyongabo, Salomey
  Osei, Ankur~P. Parikh, Laura Perez{-}Beltrachini, Niranjan~Ramesh Rao, Vikas
  Raunak, Juan~Diego Rodriguez, Sashank Santhanam, Jo{\~{a}}o Sedoc, Thibault
  Sellam, Samira Shaikh, Anastasia Shimorina, Marco Antonio~Sobrevilla
  Cabezudo, Hendrik Strobelt, Nishant Subramani, Wei Xu, Diyi Yang, Akhila
  Yerukola, and Jiawei Zhou. 2021.
\newblock \href {https://arxiv.org/abs/2102.01672} {The {GEM} benchmark:
  Natural language generation, its evaluation and metrics}.
\newblock In \emph{Proc.\ of GEM}.

\bibitem[{Gehrmann et~al.(2018)Gehrmann, Deng, and
  Rush}]{gehrmann-etal-2018-bottom}
Sebastian Gehrmann, Yuntian Deng, and Alexander Rush. 2018.
\newblock \href {https://arxiv.org/abs/1808.10792} {Bottom-up abstractive
  summarization}.
\newblock In \emph{Proc.\ of EMNLP}.

\bibitem[{Germann(2020)}]{germann-2020-university}
Ulrich Germann. 2020.
\newblock \href {https://aclanthology.org/2020.wmt-1.18} {The {U}niversity of
  {E}dinburgh{'}s submission to the {G}erman-to-{E}nglish and
  {E}nglish-to-{G}erman tracks in the {WMT} 2020 news translation and zero-shot
  translation robustness tasks}.
\newblock In \emph{Proc.\ of WMT}.

\bibitem[{Gillick and Liu(2010)}]{gillick-liu-2010-non}
Dan Gillick and Yang Liu. 2010.
\newblock \href {https://aclanthology.org/W10-0722} {Non-expert evaluation of
  summarization systems is risky}.
\newblock In \emph{Proc.\ of the {NAACL} {HLT} 2010 Workshop on Creating Speech
  and Language Data with {A}mazon{'}s Mechanical Turk}.

\bibitem[{Gleason et~al.(2020)Gleason, Pavel, McCamey, Low, Carrington, Kitani,
  and Bigham}]{GleasonCHI20}
Cole Gleason, Amy Pavel, Emma McCamey, Christina Low, Patrick Carrington,
  Kris~M. Kitani, and Jeffrey~P. Bigham. 2020.
\newblock \href
  {https://www.cs.cmu.edu/~jbigham/pubs/pdfs/2020/twitter-a11y.pdf} {Twitter
  a11y: {A} browser extension to make twitter images accessible}.
\newblock In \emph{Proc.\ of CHI}.

\bibitem[{Guo et~al.(2018)Guo, Pasunuru, and Bansal}]{guo-etal-2018-soft}
Han Guo, Ramakanth Pasunuru, and Mohit Bansal. 2018.
\newblock \href {https://arxiv.org/abs/1805.11004} {Soft layer-specific
  multi-task summarization with entailment and question generation}.
\newblock In \emph{Proc.\ of ACL}.

\bibitem[{Gwinnup and Anderson(2020)}]{gwinnup-anderson-2020-afrl}
Jeremy Gwinnup and Tim Anderson. 2020.
\newblock \href {https://aclanthology.org/2020.wmt-1.20} {The {AFRL} {WMT}20
  news translation systems}.
\newblock In \emph{Proc.\ of WMT}.

\bibitem[{Hassan et~al.(2018)Hassan, Aue, Chen, Chowdhary, Clark, Federmann,
  Huang, Junczys-Dowmunt, Lewis, Li, Liu, Liu, Luo, Menezes, Qin, Seide, Tan,
  Tian, Wu, Wu, Xia, Zhang, Zhang, and Zhou}]{Hassan2018AchievingHP}
Hany Hassan, Anthony Aue, Chang Chen, Vishal Chowdhary, Jonathan Clark,
  Christian Federmann, Xuedong Huang, Marcin Junczys-Dowmunt, William Lewis,
  Mengnan Li, Shujie Liu, Tie-Yan Liu, Renqian Luo, Arul Menezes, Tao Qin,
  Frank Seide, Xu~Tan, Fei Tian, Lijun Wu, Shuangzhi Wu, Yingce Xia, Dongdong
  Zhang, Zhirui Zhang, and Ming Zhou. 2018.
\newblock \href {https://arxiv.org/abs/1803.05567} {Achieving human parity on
  automatic {Chinese} to {English} news translation}.

\bibitem[{Heafield et~al.(2020)Heafield, Hayashi, Oda, Konstas, Finch, Neubig,
  Li, and Birch}]{wngt2020}
Kenneth Heafield, Hiroaki Hayashi, Yusuke Oda, Ioannis Konstas, Andrew Finch,
  Graham Neubig, Xian Li, and Alexandra Birch. 2020.
\newblock \href {https://aclanthology.org/2020.ngt-1.1} {Findings of the fourth
  workshop on neural generation and translation}.
\newblock In \emph{Proc.\ of WNGT}.

\bibitem[{Hermann et~al.(2015)Hermann, Kocisk{\'{y}}, Grefenstette, Espeholt,
  Kay, Suleyman, and Blunsom}]{cnndaily}
Karl~Moritz Hermann, Tom{\'{a}}s Kocisk{\'{y}}, Edward Grefenstette, Lasse
  Espeholt, Will Kay, Mustafa Suleyman, and Phil Blunsom. 2015.
\newblock \href {http://arxiv.org/abs/1506.03340} {Teaching machines to read
  and comprehend}.
\newblock In \emph{Proc.\ of NeurIPS}.

\bibitem[{Hessel et~al.(2021)Hessel, Holtzman, Forbes, Bras, and
  Choi}]{clipscore}
Jack Hessel, Ari Holtzman, Maxwell Forbes, Ronan~Le Bras, and Yejin Choi. 2021.
\newblock \href {https://arxiv.org/abs/2104.08718} {{CLIPScore}: {A}
  reference-free evaluation metric for image captioning}.
\newblock In \emph{Proc.\ of EMNLP}.

\bibitem[{Hsu et~al.(2018)Hsu, Lin, Lee, Min, Tang, and
  Sun}]{hsu-etal-2018-unified}
Wan-Ting Hsu, Chieh-Kai Lin, Ming-Ying Lee, Kerui Min, Jing Tang, and Min Sun.
  2018.
\newblock \href {https://arxiv.org/abs/1805.06266} {A unified model for
  extractive and abstractive summarization using inconsistency loss}.
\newblock In \emph{Proc.\ of ACL}.

\bibitem[{Jiang and Bansal(2018)}]{jiang-bansal-2018-closed}
Yichen Jiang and Mohit Bansal. 2018.
\newblock \href {https://arxiv.org/abs/1809.04585} {Closed-book training to
  improve summarization encoder memory}.
\newblock In \emph{Proc.\ of EMNLP}.

\bibitem[{Karpathy and Fei-Fei(2015)}]{Karpathy_2015_CVPR}
Andrej Karpathy and Li~Fei-Fei. 2015.
\newblock \href {https://arxiv.org/abs/1412.2306} {Deep visual-semantic
  alignments for generating image descriptions}.
\newblock In \emph{Proc.\ of CVPR}.

\bibitem[{Kasai et~al.(2021{\natexlab{a}})Kasai, Pappas, Peng, Cross, and
  Smith}]{deepshallow}
Jungo Kasai, Nikolaos Pappas, Hao Peng, James Cross, and Noah~A. Smith.
  2021{\natexlab{a}}.
\newblock \href {https://arxiv.org/abs/2006.10369} {Deep encoder, shallow
  decoder: Reevaluating non-autoregressive machine translation}.
\newblock In \emph{Proc.\ of ICLR}.

\bibitem[{Kasai et~al.(2021{\natexlab{b}})Kasai, Peng, Zhang, Yogatama,
  Ilharco, Pappas, Mao, Chen, and Smith}]{kasai2021t2r}
Jungo Kasai, Hao Peng, Yizhe Zhang, Dani Yogatama, Gabriel Ilharco, Nikolaos
  Pappas, Yi~Mao, Weizhu Chen, and Noah~A. Smith. 2021{\natexlab{b}}.
\newblock \href {https://arxiv.org/abs/2103.13076} {Finetuning pretrained
  transformers into {RNN}s}.
\newblock In \emph{Proc. of EMNLP}.

\bibitem[{Kasai et~al.(2022)Kasai, Sakaguchi, Dunagan, Morrison, Bras, Choi,
  and Smith}]{kasai2021thumb}
Jungo Kasai, Keisuke Sakaguchi, Lavinia Dunagan, Jacob Morrison, Ronan~Le Bras,
  Yejin Choi, and Noah~A. Smith. 2022.
\newblock \href {https://arxiv.org/abs/2111.08940} {Transparent human
  evaluation for image captioning}.
\newblock In \emph{Proc.\ of NAACL}.

\bibitem[{Khashabi et~al.(2021)Khashabi, Stanovsky, Bragg, Lourie, Kasai, Choi,
  Smith, and Weld}]{genie}
Daniel Khashabi, Gabriel Stanovsky, Jonathan Bragg, Nicholas Lourie, Jungo
  Kasai, Yejin Choi, Noah~A. Smith, and Daniel~S. Weld. 2021.
\newblock \href {https://arxiv.org/abs/2101.06561} {{GENIE:} {A} leaderboard
  for human-in-the-loop evaluation of text generation}.

\bibitem[{Kiyono et~al.(2020)Kiyono, Ito, Konno, Morishita, and
  Suzuki}]{kiyono-etal-2020-tohoku}
Shun Kiyono, Takumi Ito, Ryuto Konno, Makoto Morishita, and Jun Suzuki. 2020.
\newblock \href {https://aclanthology.org/2020.wmt-1.12} {Tohoku-{AIP}-{NTT} at
  {WMT} 2020 news translation task}.
\newblock In \emph{Proc. of WMT}.

\bibitem[{Koehn et~al.(2007)Koehn, Hoang, Birch, Callison-Burch, Federico,
  Bertoldi, Cowan, Shen, Moran, Zens, Dyer, Bojar, Constantin, and
  Herbst}]{koehn-etal-2007-moses}
Philipp Koehn, Hieu Hoang, Alexandra Birch, Chris Callison-Burch, Marcello
  Federico, Nicola Bertoldi, Brooke Cowan, Wade Shen, Christine Moran, Richard
  Zens, Chris Dyer, Ond{\v{r}}ej Bojar, Alexandra Constantin, and Evan Herbst.
  2007.
\newblock \href {https://aclanthology.org/P07-2045} {{M}oses: Open source
  toolkit for statistical machine translation}.
\newblock In \emph{Proc.\ of ACL Demo and Poster Sessions}.

\bibitem[{Kryscinski et~al.(2019)Kryscinski, Keskar, McCann, Xiong, and
  Socher}]{kryscinski-etal-2019-neural}
Wojciech Kryscinski, Nitish~Shirish Keskar, Bryan McCann, Caiming Xiong, and
  Richard Socher. 2019.
\newblock \href {https://arxiv.org/abs/1908.08960} {Neural text summarization:
  A critical evaluation}.
\newblock In \emph{Proc.\ of EMNLP}.

\bibitem[{Kry{\'s}ci{\'n}ski et~al.(2018)Kry{\'s}ci{\'n}ski, Paulus, Xiong, and
  Socher}]{kryscinski-etal-2018-improving}
Wojciech Kry{\'s}ci{\'n}ski, Romain Paulus, Caiming Xiong, and Richard Socher.
  2018.
\newblock \href {https://arxiv.org/abs/1808.07913} {Improving abstraction in
  text summarization}.
\newblock In \emph{Proc.\ of EMNLP}.

\bibitem[{Lewis et~al.(2020)Lewis, Liu, Goyal, Ghazvininejad, Mohamed, Levy,
  Stoyanov, and Zettlemoyer}]{lewis-etal-2020-bart}
Mike Lewis, Yinhan Liu, Naman Goyal, Marjan Ghazvininejad, Abdelrahman Mohamed,
  Omer Levy, Veselin Stoyanov, and Luke Zettlemoyer. 2020.
\newblock \href {https://arxiv.org/abs/1910.13461} {{BART}: Denoising
  sequence-to-sequence pre-training for natural language generation,
  translation, and comprehension}.
\newblock In \emph{Proc.\ of ACL}.

\bibitem[{Li et~al.(2020)Li, Zhao, Wang, Chen, Utiyama, and
  Sumita}]{li-etal-2020-sjtu}
Zuchao Li, Hai Zhao, Rui Wang, Kehai Chen, Masao Utiyama, and Eiichiro Sumita.
  2020.
\newblock \href {https://arxiv.org/abs/2010.05122} {{SJTU}-{NICT}{'}s
  supervised and unsupervised neural machine translation systems for the
  {WMT}20 news translation task}.
\newblock In \emph{Proc.\ of WMT}.

\bibitem[{Lin(2004)}]{Lin2004ROUGEAP}
Chin-Yew Lin. 2004.
\newblock \href {https://www.aclweb.org/anthology/W04-1013/} {{ROUGE}: A
  package for automatic evaluation of summaries}.
\newblock In \emph{Proc.\ of Text Summarization Branches Out}.

\bibitem[{Lin et~al.(2014)Lin, Maire, Belongie, Bourdev, Girshick, Hays,
  Perona, Ramanan, Doll{\'{a}}r, and Zitnick}]{mscoco}
Tsung{-}Yi Lin, Michael Maire, Serge~J. Belongie, Lubomir~D. Bourdev, Ross~B.
  Girshick, James Hays, Pietro Perona, Deva Ramanan, Piotr Doll{\'{a}}r, and
  C.~Lawrence Zitnick. 2014.
\newblock \href {http://arxiv.org/abs/1405.0312} {Microsoft {COCO:} common
  objects in context}.
\newblock In \emph{Proc.\ of ECCV}.

\bibitem[{Liu and Lapata(2019)}]{liu-lapata-2019-text}
Yang Liu and Mirella Lapata. 2019.
\newblock \href {https://arxiv.org/abs/1908.08345} {Text summarization with
  pretrained encoders}.
\newblock In \emph{Proc.\ of EMNLP}.

\bibitem[{Ma et~al.(2019)Ma, Wei, Bojar, and Graham}]{ma-etal-2019-results}
Qingsong Ma, Johnny Wei, Ond{\v{r}}ej Bojar, and Yvette Graham. 2019.
\newblock \href {https://www.aclweb.org/anthology/W19-5302} {Results of the
  {WMT}19 metrics shared task: Segment-level and strong {MT} systems pose big
  challenges}.
\newblock In \emph{Proc.\ of WMT}.

\bibitem[{Mach{\'a}{\v{c}}ek and Bojar(2014)}]{machacek-bojar-2014-results}
Matou{\v{s}} Mach{\'a}{\v{c}}ek and Ond{\v{r}}ej Bojar. 2014.
\newblock \href {https://aclanthology.org/W14-3336} {Results of the {WMT}14
  metrics shared task}.
\newblock In \emph{Proc.\ of WMT}.

\bibitem[{Manning et~al.(2014)Manning, Surdeanu, Bauer, Finkel, Bethard, and
  McClosky}]{manning-EtAl:2014:P14-5}
Christopher~D. Manning, Mihai Surdeanu, John Bauer, Jenny Finkel, Steven~J.
  Bethard, and David McClosky. 2014.
\newblock \href {http://www.aclweb.org/anthology/P/P14/P14-5010} {The
  {Stanford} {CoreNLP} natural language processing toolkit}.
\newblock In \emph{Proc.\ of ACL System Demonstrations}.

\bibitem[{Marie et~al.(2021)Marie, Fujita, and
  Rubino}]{marie-etal-2021-scientific}
Benjamin Marie, Atsushi Fujita, and Raphael Rubino. 2021.
\newblock \href {https://arxiv.org/abs/2106.15195} {Scientific credibility of
  machine translation research: A meta-evaluation of 769 papers}.
\newblock In \emph{Proc.\ of ACL}.

\bibitem[{Mathur et~al.(2020{\natexlab{a}})Mathur, Baldwin, and
  Cohn}]{tangledup}
Nitika Mathur, Timothy Baldwin, and Trevor Cohn. 2020{\natexlab{a}}.
\newblock \href {https://arxiv.org/abs/2006.06264} {Tangled up in {BLEU}:
  Reevaluating the evaluation of automatic machine translation evaluation
  metrics}.
\newblock In \emph{Proc.\ of ACL}.

\bibitem[{Mathur et~al.(2020{\natexlab{b}})Mathur, Wei, Freitag, Ma, and
  Bojar}]{mathur-etal-2020-results}
Nitika Mathur, Johnny Wei, Markus Freitag, Qingsong Ma, and Ond{\v{r}}ej Bojar.
  2020{\natexlab{b}}.
\newblock \href {https://aclanthology.org/2020.wmt-1.77} {Results of the
  {WMT}20 metrics shared task}.
\newblock In \emph{Proc.\ of WMT}.

\bibitem[{Meng et~al.(2020)Meng, Yan, Liu, Gao, Zeng, Zeng, Li, Chen, Zhou,
  Liu, and Zhou}]{meng-etal-2020-wechat}
Fandong Meng, Jianhao Yan, Yijin Liu, Yuan Gao, Xianfeng Zeng, Qinsong Zeng,
  Peng Li, Ming Chen, Jie Zhou, Sifan Liu, and Hao Zhou. 2020.
\newblock \href {https://arxiv.org/abs/2010.00247} {{W}e{C}hat neural machine
  translation systems for {WMT}20}.
\newblock In \emph{Proc.\ of WMT}.

\bibitem[{Mishra and Arunkumar(2021)}]{MishraA21}
Swaroop Mishra and Anjana Arunkumar. 2021.
\newblock \href {https://arxiv.org/abs/2106.05532} {How robust are model
  rankings : {A} leaderboard customization approach for equitable evaluation}.
\newblock In \emph{Proc.\ of AAAI}.

\bibitem[{Molchanov(2020)}]{molchanov-2020-promt}
Alexander Molchanov. 2020.
\newblock \href {https://aclanthology.org/2020.wmt-1.25} {{PROMT} systems for
  {WMT} 2020 shared news translation task}.
\newblock In \emph{Proc.\ of WMT}.

\bibitem[{Nallapati et~al.(2016)Nallapati, Zhou, dos Santos,
  G{\"{u}}l{\c{c}}ehre, and Xiang}]{nallapati-etal-2016-abstractive}
Ramesh Nallapati, Bowen Zhou, C{\'{\i}}cero~Nogueira dos Santos, {\c{C}}aglar
  G{\"{u}}l{\c{c}}ehre, and Bing Xiang. 2016.
\newblock \href {https://arxiv.org/abs/1602.06023} {Abstractive text
  summarization using sequence-to-sequence {RNN}s and beyond}.
\newblock In \emph{Proc.\ of CoNLL}.

\bibitem[{Narayan et~al.(2018)Narayan, Cohen, and
  Lapata}]{narayan-etal-2018-ranking}
Shashi Narayan, Shay~B. Cohen, and Mirella Lapata. 2018.
\newblock \href {https://arxiv.org/abs/1802.08636} {Ranking sentences for
  extractive summarization with reinforcement learning}.
\newblock In \emph{Proc.\ of NAACL}.

\bibitem[{Ng et~al.(2019)Ng, Yee, Baevski, Ott, Auli, and
  Edunov}]{ng-etal-2019-facebook}
Nathan Ng, Kyra Yee, Alexei Baevski, Myle Ott, Michael Auli, and Sergey Edunov.
  2019.
\newblock \href {https://arxiv.org/abs/1907.06616} {{F}acebook {FAIR}{'}s
  {WMT}19 news translation task submission}.
\newblock In \emph{Proc.\ of WMT}.

\bibitem[{Oravecz et~al.(2020)Oravecz, Bontcheva, Tihanyi, Kolovratnik,
  Bhaskar, Lardilleux, Klocek, and Eisele}]{oravecz-etal-2020-etranslations}
Csaba Oravecz, Katina Bontcheva, L{\'a}szl{\'o} Tihanyi, David Kolovratnik,
  Bhavani Bhaskar, Adrien Lardilleux, Szymon Klocek, and Andreas Eisele. 2020.
\newblock \href {https://aclanthology.org/2020.wmt-1.26} {e{T}ranslation{'}s
  submissions to the {WMT} 2020 news translation task}.
\newblock In \emph{Proc.\ of WMT}.

\bibitem[{Ott et~al.(2019)Ott, Edunov, Baevski, Fan, Gross, Ng, Grangier, and
  Auli}]{ott-etal-2019-fairseq}
Myle Ott, Sergey Edunov, Alexei Baevski, Angela Fan, Sam Gross, Nathan Ng,
  David Grangier, and Michael Auli. 2019.
\newblock \href {https://arxiv.org/abs/1904.01038} {fairseq: A fast, extensible
  toolkit for sequence modeling}.
\newblock In \emph{Proc.\ of NAACL Demonstrations}.

\bibitem[{Papineni et~al.(2002)Papineni, Roukos, Ward, and
  Zhu}]{Papineni2001BleuAM}
Kishore Papineni, Salim Roukos, Todd Ward, and Wei-Jing Zhu. 2002.
\newblock \href {https://www.aclweb.org/anthology/P02-1040} {{BLEU}: a method
  for automatic evaluation of machine translation}.
\newblock In \emph{Proc.\ of ACL}.

\bibitem[{Pasunuru and Bansal(2018)}]{pasunuru-bansal-2018-multi}
Ramakanth Pasunuru and Mohit Bansal. 2018.
\newblock \href {https://arxiv.org/abs/1804.06451} {Multi-reward reinforced
  summarization with saliency and entailment}.
\newblock In \emph{Proc.\ of NAACL}.

\bibitem[{Peng et~al.(2021)Peng, Pappas, Yogatama, Schwartz, Smith, and
  Kong}]{peng2021rfa}
Hao Peng, Nikolaos Pappas, Dani Yogatama, Roy Schwartz, Noah~A.\ Smith, and
  Lingpeng Kong. 2021.
\newblock \href {https://arxiv.org/abs/2103.02143} {Random feature attention}.
\newblock In \emph{Proc. of ICLR}.

\bibitem[{Popovi{\'c}(2015)}]{popovic-2015-chrf}
Maja Popovi{\'c}. 2015.
\newblock \href {https://aclanthology.org/W15-3049/} {chr{F}: character n-gram
  {F}-score for automatic {MT} evaluation}.
\newblock In \emph{Proc.\ of WMT}.

\bibitem[{Popovi{\'c}(2017)}]{popovic-2017-chrf}
Maja Popovi{\'c}. 2017.
\newblock \href {https://aclanthology.org/W17-4770} {chr{F}++: words helping
  character n-grams}.
\newblock In \emph{Proc.\ of WMT}.

\bibitem[{Post(2018)}]{post-2018-call}
Matt Post. 2018.
\newblock \href {https://www.aclweb.org/anthology/W18-6319} {A call for clarity
  in reporting {BLEU} scores}.
\newblock In \emph{Proc.\ of WMT}.

\bibitem[{Radford et~al.(2021)Radford, Kim, Hallacy, Ramesh, Goh, Agarwal,
  Sastry, Askell, Mishkin, Clark, Krueger, and Sutskever}]{clip2021}
Alec Radford, Jong~Wook Kim, Chris Hallacy, Aditya Ramesh, Gabriel Goh,
  Sandhini Agarwal, Girish Sastry, Amanda Askell, Pamela Mishkin, Jack Clark,
  Gretchen Krueger, and Ilya Sutskever. 2021.
\newblock \href {https://arxiv.org/abs/2103.00020} {Learning transferable
  visual models from natural language supervision}.

\bibitem[{Raffel et~al.(2020)Raffel, Shazeer, Roberts, Lee, Narang, Matena,
  Zhou, Li, and Liu}]{T5}
Colin Raffel, Noam Shazeer, Adam Roberts, Katherine Lee, Sharan Narang, Michael
  Matena, Yanqi Zhou, Wei Li, and Peter~J. Liu. 2020.
\newblock \href {http://jmlr.org/papers/v21/20-074.html} {Exploring the limits
  of transfer learning with a unified text-to-text transformer}.
\newblock \emph{JLMR}.

\bibitem[{Rajpurkar et~al.(2016)Rajpurkar, Zhang, Lopyrev, and
  Liang}]{rajpurkar-etal-2016-squad}
Pranav Rajpurkar, Jian Zhang, Konstantin Lopyrev, and Percy Liang. 2016.
\newblock \href {https://arxiv.org/abs/1606.05250} {{SQ}u{AD}: 100,000+
  questions for machine comprehension of text}.
\newblock In \emph{Proc.\ of EMNLP}.

\bibitem[{Rei et~al.(2020)Rei, Stewart, Farinha, and
  Lavie}]{rei-etal-2020-comet}
Ricardo Rei, Craig Stewart, Ana~C Farinha, and Alon Lavie. 2020.
\newblock \href {https://arxiv.org/abs/2009.09025} {{COMET}: A neural framework
  for {MT} evaluation}.
\newblock In \emph{Proc.\ of EMNLP}.

\bibitem[{Russakovsky et~al.(2015)Russakovsky, Deng, Su, Krause, Satheesh, Ma,
  Huang, Karpathy, Khosla, Bernstein, Berg, and Fei-Fei}]{imagenet-challenge}
Olga Russakovsky, Jia Deng, Hao Su, Jonathan Krause, Sanjeev Satheesh, Sean Ma,
  Zhiheng Huang, Andrej Karpathy, Aditya Khosla, Michael Bernstein,
  Alexander~C. Berg, and Li~Fei-Fei. 2015.
\newblock \href {https://arxiv.org/abs/1409.0575} {{ImageNet} large scale
  visual recognition challenge}.
\newblock \emph{IJCV}.

\bibitem[{Schwartz et~al.(2019)Schwartz, Dodge, Smith, and Etzioni}]{greenai}
Roy Schwartz, Jesse Dodge, Noah~A. Smith, and Oren Etzioni. 2019.
\newblock \href {http://arxiv.org/abs/1907.10597} {Green {AI}}.
\newblock \emph{CACM}.

\bibitem[{Scialom et~al.(2019)Scialom, Lamprier, Piwowarski, and
  Staiano}]{scialom-etal-2019-answers}
Thomas Scialom, Sylvain Lamprier, Benjamin Piwowarski, and Jacopo Staiano.
  2019.
\newblock \href {https://arxiv.org/abs/1909.01610} {Answers unite! unsupervised
  metrics for reinforced summarization models}.
\newblock In \emph{Proc.\ of EMNLP}.

\bibitem[{See et~al.(2017)See, Liu, and Manning}]{see-etal-2017-get}
Abigail See, Peter~J. Liu, and Christopher~D. Manning. 2017.
\newblock \href {https://arxiv.org/abs/1704.04368} {Get to the point:
  Summarization with pointer-generator networks}.
\newblock In \emph{Proc.\ of ACL}.

\bibitem[{Sellam et~al.(2020)Sellam, Das, and Parikh}]{sellam2020bleurt}
Thibault Sellam, Dipanjan Das, and Ankur~P Parikh. 2020.
\newblock \href {https://arxiv.org/abs/2004.04696} {{BLEURT}: Learning robust
  metrics for text generation}.
\newblock In \emph{Proc.\ of ACL}.

\bibitem[{Sennrich et~al.(2016{\natexlab{a}})Sennrich, Haddow, and
  Birch}]{sennrich-etal-2016-improving}
Rico Sennrich, Barry Haddow, and Alexandra Birch. 2016{\natexlab{a}}.
\newblock \href {https://arxiv.org/abs/1511.06709} {Improving neural machine
  translation models with monolingual data}.
\newblock In \emph{Proc.\ of ACL}.

\bibitem[{Sennrich et~al.(2016{\natexlab{b}})Sennrich, Haddow, and
  Birch}]{sennrich-etal-2016-neural}
Rico Sennrich, Barry Haddow, and Alexandra Birch. 2016{\natexlab{b}}.
\newblock \href {https://www.aclweb.org/anthology/P16-1162} {Neural machine
  translation of rare words with subword units}.
\newblock In \emph{Proc.\ of ACL}.

\bibitem[{Sharma et~al.(2019)Sharma, Huang, Hu, and
  Wang}]{sharma-etal-2019-entity}
Eva Sharma, Luyang Huang, Zhe Hu, and Lu~Wang. 2019.
\newblock \href {https://arxiv.org/abs/1909.02059} {An entity-driven framework
  for abstractive summarization}.
\newblock In \emph{Proc.\ of EMNLP}.

\bibitem[{Shi et~al.(2020)Shi, Zhao, Li, Wang, Zhang, Ai, Dang, Zhengshan, and
  Hao}]{shi-etal-2020-oppos}
Tingxun Shi, Shiyu Zhao, Xiaopu Li, Xiaoxue Wang, Qian Zhang, Di~Ai, Dawei
  Dang, Xue Zhengshan, and Jie Hao. 2020.
\newblock \href {https://aclanthology.org/2020.wmt-1.30} {{OPPO}{'}s machine
  translation systems for {WMT}20}.
\newblock In \emph{Proc.\ of WMT}.

\bibitem[{Snover et~al.(2006)Snover, Dorr, Schwartz, Micciulla, and
  Makhoul}]{TER2006}
Matthew Snover, Bonnie Dorr, Rich Schwartz, Linnea Micciulla, and John Makhoul.
  2006.
\newblock \href {https://aclanthology.org/2006.amta-papers.25} {A study of
  translation edit rate with targeted human annotation}.
\newblock In \emph{Proc.\ of AMTA}.

\bibitem[{Stanovsky et~al.(2019)Stanovsky, Smith, and
  Zettlemoyer}]{stanovsky-etal-2019-evaluating}
Gabriel Stanovsky, Noah~A. Smith, and Luke Zettlemoyer. 2019.
\newblock \href {https://arxiv.org/abs/1906.00591} {Evaluating gender bias in
  machine translation}.
\newblock In \emph{Proc.\ of ACL}.

\bibitem[{Thompson and Post(2020)}]{thompson-post-2020-automatic}
Brian Thompson and Matt Post. 2020.
\newblock \href {https://arxiv.org/abs/2004.14564} {Automatic machine
  translation evaluation in many languages via zero-shot paraphrasing}.
\newblock In \emph{Proc.\ of EMNLP}.

\bibitem[{Tibshirani(1994)}]{Tibshirani94regressionshrinkage}
Robert Tibshirani. 1994.
\newblock \href {https://www.jstor.org/stable/2346178} {Regression shrinkage
  and selection via the {L}asso}.
\newblock \emph{Journal of the Royal Statistical Society, Series B}.

\bibitem[{Toral et~al.(2018)Toral, Castilho, Hu, and
  Way}]{toral-etal-2018-attaining}
Antonio Toral, Sheila Castilho, Ke~Hu, and Andy Way. 2018.
\newblock \href {https://arxiv.org/abs/1808.10432} {Attaining the unattainable?
  reassessing claims of human parity in neural machine translation}.
\newblock In \emph{Proc.\ of WMT}.

\bibitem[{Vaswani et~al.(2017)Vaswani, Shazeer, Parmar, Uszkoreit, Jones,
  Gomez, Kaiser, and Polosukhin}]{Vaswani2017AttentionIA}
Ashish Vaswani, Noam Shazeer, Niki Parmar, Jakob Uszkoreit, Llion Jones,
  Aidan~N. Gomez, \L{}ukasz Kaiser, and Illia Polosukhin. 2017.
\newblock \href {https://arxiv.org/pdf/1706.03762.pdf} {Attention is all you
  need}.
\newblock In \emph{Proc. of NeurIPS}.

\bibitem[{Vedantam et~al.(2015)Vedantam, Zitnick, and Parikh}]{cider2015}
Ramakrishna Vedantam, C.~Lawrence Zitnick, and Devi Parikh. 2015.
\newblock \href {https://arxiv.org/abs/1411.5726} {{CIDEr}: Consensus-based
  image description evaluation}.
\newblock In \emph{Proc.\ of CVPR}.

\bibitem[{Voita et~al.(2019)Voita, Sennrich, and Titov}]{voita-etal-2019-good}
Elena Voita, Rico Sennrich, and Ivan Titov. 2019.
\newblock \href {https://arxiv.org/abs/1905.05979} {When a good translation is
  wrong in context: Context-aware machine translation improves on deixis,
  ellipsis, and lexical cohesion}.
\newblock In \emph{Proc.\ of ACL}.

\bibitem[{Wang et~al.(2016)Wang, Peter, Rosendahl, and
  Ney}]{wang-etal-2016-character}
Weiyue Wang, Jan-Thorsten Peter, Hendrik Rosendahl, and Hermann Ney. 2016.
\newblock \href {https://aclanthology.org/W16-2342} {{C}harac{T}er: Translation
  edit rate on character level}.
\newblock In \emph{Proc.\ of WMT}.

\bibitem[{Wei et~al.(2020)Wei, Shang, Wu, Yu, Li, Guo, Wang, Yang, Lei, Qin,
  and Sun}]{wei-etal-2020-hw}
Daimeng Wei, Hengchao Shang, Zhanglin Wu, Zhengzhe Yu, Liangyou Li, Jiaxin Guo,
  Minghan Wang, Hao Yang, Lizhi Lei, Ying Qin, and Shiliang Sun. 2020.
\newblock \href {https://aclanthology.org/2020.wmt-1.31} {{HW}-{TSC}{'}s
  participation in the {WMT} 2020 news translation shared task}.
\newblock In \emph{Proc.\ of WMT}.

\bibitem[{Wu et~al.(2020{\natexlab{a}})Wu, Pan, Lin, Zhu, Wang, and
  Li}]{wu-etal-2020-volctrans}
Liwei Wu, Xiao Pan, Zehui Lin, Yaoming Zhu, Mingxuan Wang, and Lei Li.
  2020{\natexlab{a}}.
\newblock \href {https://arxiv.org/abs/2010.14806} {The {V}olctrans machine
  translation system for {WMT}20}.
\newblock In \emph{Proc.\ of WMT}.

\bibitem[{Wu et~al.(2020{\natexlab{b}})Wu, Wang, Wang, Liu, Xie, Tu, Shi, and
  Li}]{wu-etal-2020-tencent}
Shuangzhi Wu, Xing Wang, Longyue Wang, Fangxu Liu, Jun Xie, Zhaopeng Tu,
  Shuming Shi, and Mu~Li. 2020{\natexlab{b}}.
\newblock \href {https://aclanthology.org/2020.wmt-1.34} {Tencent neural
  machine translation systems for the {WMT}20 news translation task}.
\newblock In \emph{Proc.\ of WMT}.

\bibitem[{Wu and Hu(2018)}]{rnes18}
Yuxiang Wu and Baotian Hu. 2018.
\newblock \href {https://arxiv.org/abs/1804.07036} {Learning to extract
  coherent summary via deep reinforcement learning}.
\newblock In \emph{Proc.\ of AAAI}.

\bibitem[{Xu and Durrett(2019)}]{xu-durrett-2019-neural}
Jiacheng Xu and Greg Durrett. 2019.
\newblock \href {https://arxiv.org/abs/1902.00863} {Neural extractive text
  summarization with syntactic compression}.
\newblock In \emph{Proc.\ of EMNLP}.

\bibitem[{Yu et~al.(2020)Yu, Sartran, Huang, Stokowiec, Donato, Srinivasan,
  Andreev, Ling, Mokra, Dal~Lago, Doron, Young, Blunsom, and
  Dyer}]{yu-etal-2020-deepmind}
Lei Yu, Laurent Sartran, Po-Sen Huang, Wojciech Stokowiec, Domenic Donato,
  Srivatsan Srinivasan, Alek Andreev, Wang Ling, Sona Mokra, Agustin Dal~Lago,
  Yotam Doron, Susannah Young, Phil Blunsom, and Chris Dyer. 2020.
\newblock \href {https://aclanthology.org/2020.wmt-1.36} {The {D}eep{M}ind
  {C}hinese{--}{E}nglish document translation system at {WMT}2020}.
\newblock In \emph{Proc.\ of WMT}.

\bibitem[{Zhang et~al.(2020{\natexlab{a}})Zhang, Zhao, Saleh, and
  Liu}]{zhang2019pegasus}
Jingqing Zhang, Yao Zhao, Mohammad Saleh, and Peter~J. Liu. 2020{\natexlab{a}}.
\newblock \href {https://arxiv.org/abs/1912.08777} {{PEGASUS}: Pre-training
  with extracted gap-sentences for abstractive summarization}.
\newblock In \emph{Proc.\ of ICML}.

\bibitem[{Zhang et~al.(2021)Zhang, Li, Hu, Yang, Zhang, Wang, Choi, and
  Gao}]{zhang2021vinvl}
Pengchuan Zhang, Xiujun Li, Xiaowei Hu, Jianwei Yang, Lei Zhang, Lijuan Wang,
  Yejin Choi, and Jianfeng Gao. 2021.
\newblock \href {https://arxiv.org/abs/2101.00529} {{VinVL}: Making visual
  representations matter in vision-language models}.
\newblock In \emph{Proc.\ of CVPR}.

\bibitem[{Zhang et~al.(2020{\natexlab{b}})Zhang, Kishore, Wu, Weinberger, and
  Artzi}]{bertscore}
Tianyi Zhang, Varsha Kishore, Felix Wu, Kilian~Q. Weinberger, and Yoav Artzi.
  2020{\natexlab{b}}.
\newblock \href {https://arxiv.org/abs/1904.09675} {{BERTScore}: Evaluating
  text generation with {BERT}}.
\newblock In \emph{Proc.\ of ICLR}.

\bibitem[{Zhang et~al.(2018)Zhang, Lapata, Wei, and
  Zhou}]{zhang-etal-2018-neural}
Xingxing Zhang, Mirella Lapata, Furu Wei, and Ming Zhou. 2018.
\newblock \href {https://arxiv.org/abs/1808.07187} {Neural latent extractive
  document summarization}.
\newblock In \emph{Proc.\ of EMNLP}.

\bibitem[{Zhou et~al.(2020)Zhou, Palangi, Zhang, Hu, Corso, and
  Gao}]{Unified-VLP}
Luowei Zhou, Hamid Palangi, Lei Zhang, Houdong Hu, Jason~J. Corso, and Jianfeng
  Gao. 2020.
\newblock \href {https://arxiv.org/abs/1909.11059} {Unified vision-language
  pre-training for image captioning and {VQA}}.
\newblock In \emph{Proc.\ of AAAI}.

\bibitem[{Zhou et~al.(2018)Zhou, Yang, Wei, Huang, Zhou, and
  Zhao}]{zhou-etal-2018-neural-document}
Qingyu Zhou, Nan Yang, Furu Wei, Shaohan Huang, Ming Zhou, and Tiejun Zhao.
  2018.
\newblock \href {https://arxiv.org/abs/1807.02305} {Neural document
  summarization by jointly learning to score and select sentences}.
\newblock In \emph{Proc.\ of ACL}.

\bibitem[{Ziegler et~al.(2019)Ziegler, Stiennon, Wu, Brown, Radford, Amodei,
  Christiano, and Irving}]{gpt2-zero-shot-2019}
Daniel~M. Ziegler, Nisan Stiennon, Jeffrey Wu, Tom~B. Brown, Alec Radford,
  Dario Amodei, Paul~F. Christiano, and Geoffrey Irving. 2019.
\newblock \href {http://arxiv.org/abs/1909.08593} {Fine-tuning language models
  from human preferences}.

\end{thebibliography}
\bibliographystyle{acl_natbib}

\clearpage
\appendix
\begin{appendices}

\section{Case Studies of Evaluation Practice}
\label{appendix:breakdown}
\begin{figure*}[t]
\centering
    \includegraphics[height=4.27cm]{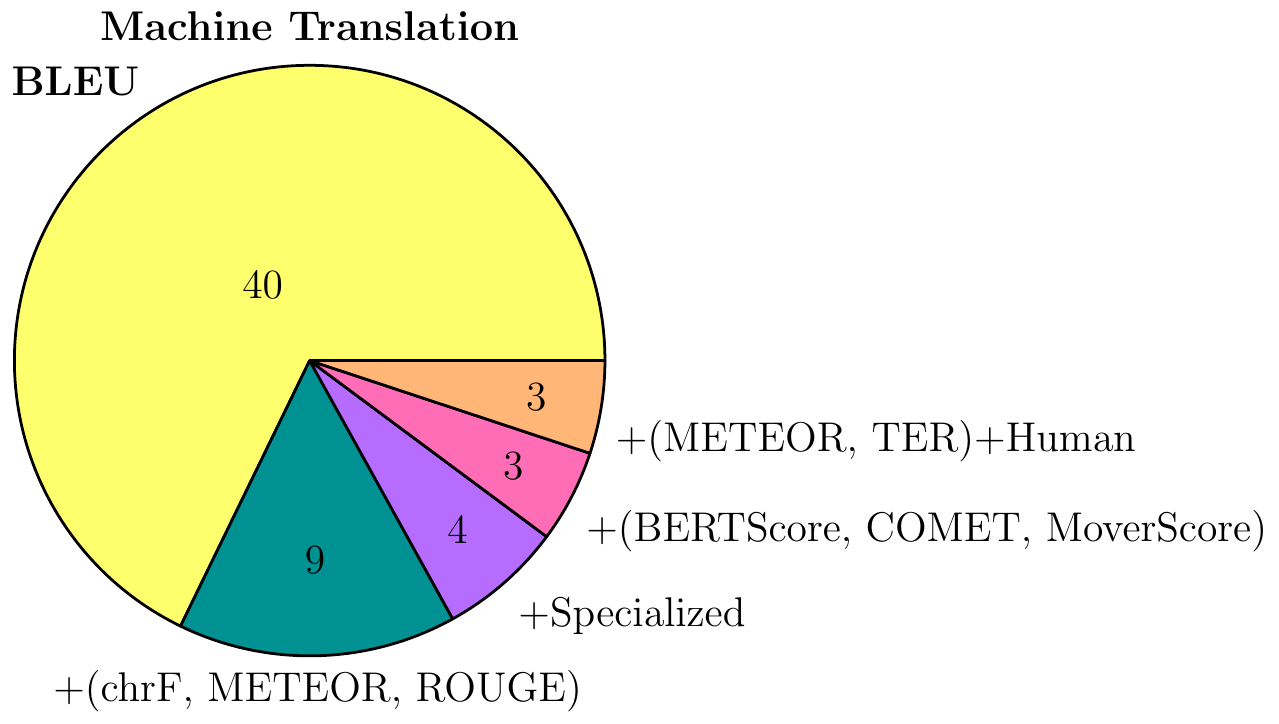}
    \includegraphics[height=4.27cm]{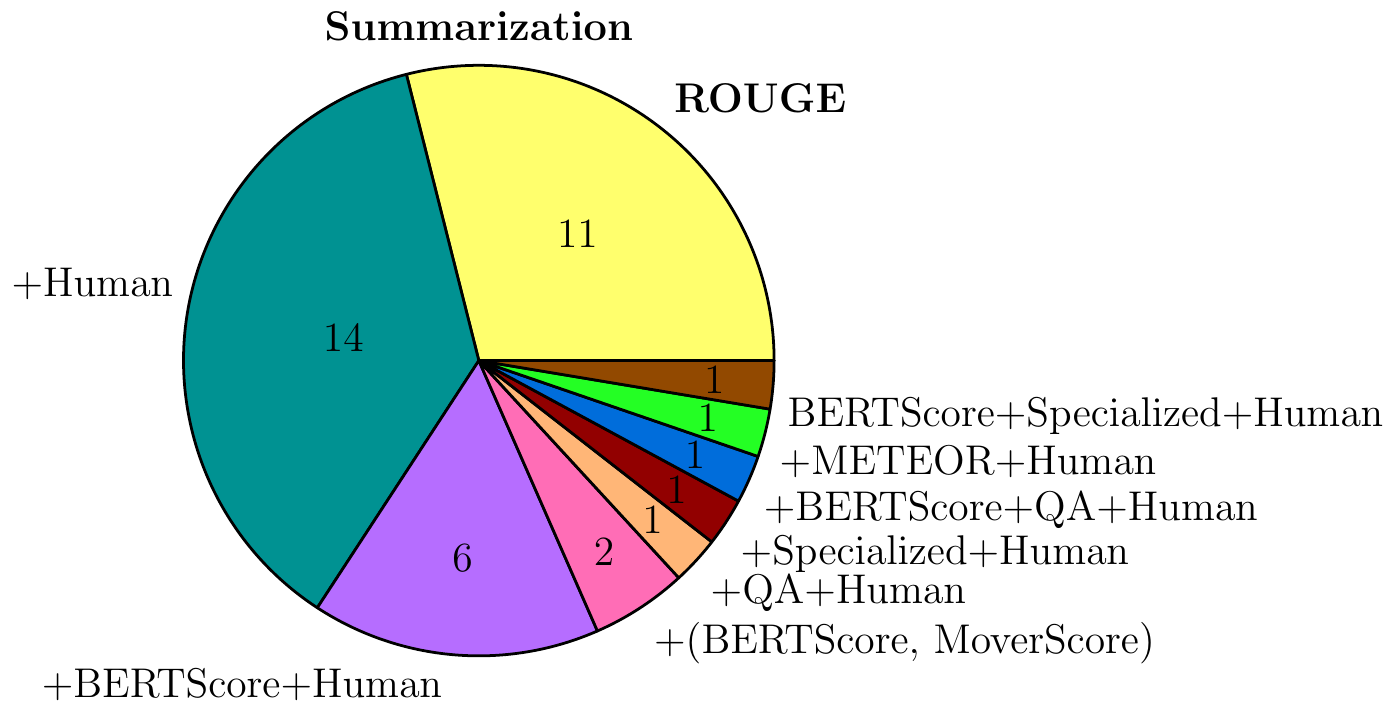}
\caption{Breakdowns of evaluation metrics used in the papers on machine translation and summarization from NAACL and ACL 2021. We examined all papers whose title contains ``machine translation'' and ``summarization'' and disregarded papers primarily on evaluation metrics. ``QA'' metrics use a QA system to evaluate summaries (e.g., \citealp{eyal-etal-2019-question}). ``Specialized'' indicates specialized evaluation in a particular dimension, rather than the overall generation quality, such as document-level evaluations on contrastive sets \cite{voita-etal-2019-good}.}
\label{fig:mt-summ-survey}
\end{figure*}

Fig.\ \ref{fig:mt-summ-survey} depicts breakdowns of evaluation metrics used in the papers on machine translation and summarization from NAACL and ACL 2021. We examined all papers whose title contains ``machine translation'' and ``summarization.'' 
We see the clear gap between generation modeling and evaluation research; most researchers do not take advantage of recent metrics that correlate better with human judgments.

\section{Participating Generators}
Here we list the generators submitted in the initial \bilboards.
\label{appendix:generators}
\subsection{WMT20 ZH-EN}
\label{appendix:wmt20-zh-en_generators}

\begin{table}[h]
\small
\centering
\begin{tabular}{@{} l@{\hspace{-0.2cm}} r @{}}
\toprule[.1em]
\textbf{Hyperparameter} & \textbf{Value}\\
\midrule[.1em]
label smoothing & 0.1\\
\# max tokens & 4096 \\
dropout rate & 0.1 \\
encoder embedding dim  & 512\\
encoder ffn dim  & 2048\\
\# encoder attn heads & 8\\
decoder embedding dim  & 512\\
decoder ffn dim  & 2048\\
\# decoder attn heads & 8\\
max source positions & 1024 \\
max target positions & 1024 \\
Adam lrate& $5\times 10^{-4}$ \\
Adam $\beta_1$& 0.9\\
Adam $\beta_2$& 0.98\\
lr-scheduler &  inverse square \\
warm-up lr & $1\times 10^{-7}$ \\
\# warmup updates & 4000 \\
\# max updates &  600K \\
\# GPUs &  8 \\
length penalty & 0.6\\
\bottomrule[.1em]
\end{tabular}
\caption{Transformer-base \texttt{fairseq} hyperparameters and setting.}
\label{tab:base-setting}
\end{table}

\begin{table}[h]
\small
\centering
\begin{tabular}{@{} l@{\hspace{-0.2cm}} r @{}}
\toprule[.1em]
\textbf{Hyperparameter} & \textbf{Value}\\
\midrule[.1em]
label smoothing & 0.1\\
\# max tokens & 4096 \\
dropout rate & 0.1\\
encoder embedding dim  & 1024\\
encoder ffn dim  & 4096\\
\# encoder attn heads & 16\\
decoder embedding dim  & 1024\\
decoder ffn dim  & 4096\\
\# decoder attn heads & 16\\
max source positions & 1024 \\
max target positions & 1024 \\
Adam lrate& $5\times 10^{-4}$ \\
Adam $\beta_1$& 0.9\\
Adam $\beta_2$& 0.98\\
lr-scheduler &  inverse square \\
warm-up lr & $1\times 10^{-7}$ \\
\# warmup updates & 4000 \\
\# max updates &  600K \\
\# GPUs &  8 \\
length penalty & 0.6\\
\bottomrule[.1em]
\end{tabular}
\caption{Transformer-large and transformer-large-ensemble \texttt{fairseq} hyperparameters and setting. Transformer-large-ensemble ensembles four transformer-large models with different random initializations.}
\label{tab:large-setting}
\end{table}

We use all 16 submissions for the WMT20 ZH-EN task \cite{barrault-etal-2020-findings}\footnote{\url{https://www.statmt.org/wmt20/results.html}.} as well as our own three transformer baselines that were implemented in \texttt{fairseq} \cite{ott-etal-2019-fairseq}.
Our baselines allow researchers to compare their translation models without resource-intensive techniques such as backtranslation \cite{sennrich-etal-2016-improving}, model ensembling, and deep encoders \cite{deepshallow}.
Tables \ref{tab:base-setting} and \ref{tab:large-setting} list the hyperprameters.
We generally follow the setting from \citet{Vaswani2017AttentionIA}.
We use \texttt{newstest-2019} as the dev.\ set and the official training data.\footnote{\url{http://www.statmt.org/wmt20/translation-task.html}.}
We apply Moses tokenization \cite{koehn-etal-2007-moses} and BPE with 32K operations \cite{sennrich-etal-2016-neural} to English text. 
We tokenize Chinese text with the Jieba package,\footnote{\url{https://github.com/fxsjy/jieba}.} following \citet{Hassan2018AchievingHP}.
Separately from English, BPE with 32K operations is then applied to Chinese.
The decoder input and output embeddings are tied.
Moses detokenization is applied to get the final outputs in the last step.
We make the three models and preprocessed train/dev.\ data publicly available.\footnote{\url{https://github.com/jungokasai/billboard/tree/master/baselines}.}
Table \ref{tab:wmt20-zh-en_generators} lists all generators and their automatic evaluation scores from the top-performing metric (ensemble in this case).

\begin{table}[h!]
\small
\centering
\begin{tabular}{@{} l l c  @{}}
\toprule[.1em]
\textbf{Generator} & \textbf{Description} & \textbf{Score} \\
\midrule[.1em]
Huoshan Translate & \citet{wu-etal-2020-volctrans}&78.85\\
THUNLP & Not available&78.81\\
Huawei TSC & \citet{wei-etal-2020-hw}&78.79 \\
DeepMind & \citet{yu-etal-2020-deepmind}&78.76\\
WeChat AI & \citet{meng-etal-2020-wechat} & 78.75\\
Tencent Translation & \citet{wu-etal-2020-tencent} & 78.74 \\
DiDi NLP & \citet{chen-etal-2020-didis} & 78.66\\
OPPO & \citet{shi-etal-2020-oppos}&78.59\\
Online-B & Not available & 78.36\\
SJTU-NICT & \citet{li-etal-2020-sjtu} & 78.27\\
trans-large-ensemble & \S\ref{appendix:wmt20-zh-en_generators} & 77.35 \\
trans-large & \S\ref{appendix:wmt20-zh-en_generators} & 76.98\\
Online-A & Not available&76.86\\
trans-base & \S\ref{appendix:wmt20-zh-en_generators} & 76.79\\
dong-nmt & Not available & 76.74\\
Online-G & Not available & 76.44\\
zlabs-nlp	 & Not available & 75.79\\
Online-Z & Not available & 75.05\\
WMT Biomed Baseline & \citet{bawden-etal-2020-findings} & 73.89\\
\bottomrule[.1em]
\end{tabular}
\caption{WMT20 ZH-EN generators and reference papers. The score column indicates the score from the metric that currently correlates best with the human judgments (\textbf{ensemble}).}
\label{tab:wmt20-zh-en_generators}
\end{table}

\subsection{WMT20 EN-DE}
\label{appendix:wmt20-en-de_generators}
Similar to WMT20 ZH-EN, we use all 14 submissions for the WMT20 EN-DE task along with our three transformer baselines.
The same hyperparameters are chosen as in WMT20 ZH-EN (Tables \ref{tab:base-setting} and \ref{tab:large-setting}). 
We preprocess both English and German text by the Moses tokenizer and \textit{joint} BPE with 32K operations. All embeddings are shared.
We apply the Moses detokenizer to get the final outputs.
Table \ref{tab:wmt20-en-de_generators} shows the generators and their automatic evaluation scores from the top-performing metric (ensemble).
\begin{table}[h!]
\small
\centering
\begin{tabular}{@{} l l c  @{}}
\toprule[.1em]
\textbf{Generator} & \textbf{Description} & \textbf{Score} \\
\midrule[.1em]
Tohoku-AIP-NTT & \citet{kiyono-etal-2020-tohoku}&90.50\\
Tencent Translate & \citet{wu-etal-2020-tencent} & 90.43\\
OPPO & \citet{shi-etal-2020-oppos}& 90.42 \\
eTranslation & \citet{oravecz-etal-2020-etranslations}& 90.39\\
Online-B & Not available & 90.38\\
Huoshan Translate & \citet{wu-etal-2020-volctrans} & 90.32 \\
\multicolumn{2}{@{}l}{AFRL \quad \quad \quad \quad \citet{gwinnup-anderson-2020-afrl} }  & 90.16\\
Online-A & Not available&90.12\\
UEDIN &\citet{germann-2020-university}&89.98\\
PROMT NMT & \citet{molchanov-2020-promt} & 89.66\\
trans-large & \S\ref{appendix:wmt20-en-de_generators} & 89.60 \\
trans-large-ensemble & \S\ref{appendix:wmt20-en-de_generators} & 89.59\\
trans-base & \S\ref{appendix:wmt20-en-de_generators} & 89.35\\
Online-Z & Not available&89.26\\
Online-G & Not available & 88.98\\
zlabs-nlp	 & Not available & 88.65\\
WMT Biomed Baseline & \citet{bawden-etal-2020-findings} & 88.23\\
\bottomrule[.1em]
\end{tabular}
\caption{WMT20 EN-DE generators and reference papers. The score column indicates the score from the metric that currently correlates best with the human judgments (\textbf{ensemble}).}
\label{tab:wmt20-en-de_generators}
\end{table}

\subsection{CNNDM Summarization}
We submit all 26 models from \citet{fabbri2021summeval}.\footnote{\url{https://github.com/Yale-LILY/SummEval}.}
Table \ref{tab:cnndm_generators} shows all models and their automatic evaluation scores from the top-performing metric (COMET).
\begin{table}[h!]
\small
\centering
\begin{tabular}{@{} l l c  @{}}
\toprule[.1em]
\textbf{Generator} & \textbf{Description} & \textbf{Score} \\
\midrule[.1em]
Lead-3 & First 3 sentences & -0.011\\
T5 & \citet{T5} & -0.030\\
BART & \citet{lewis-etal-2020-bart} & -0.032\\
Pegasus-dynamic-mix & \citet{zhang2019pegasus} & -0.044\\
RNES & \citet{rnes18} & -0.049\\
Unified-ext-abs & \citet{hsu-etal-2018-unified} & -0.056\\
Pegasus-huge-news & \citet{zhang2019pegasus} & -0.056\\
REFRESH & \citet{narayan-etal-2018-ranking} & -0.067\\
\multicolumn{2}{@{}l}{ROUGESal \quad \quad \quad \quad \citet{pasunuru-bansal-2018-multi}} & -0.073\\
Human-H & Highlights & -0.075\\
NEUSUM & \citet{zhou-etal-2018-neural-document} & -0.083 \\
BanditSum & \citet{dong-etal-2018-banditsum} & -0.083\\
LATENT & \citet{zhang-etal-2018-neural} & -0.099\\
Closed-book-decoder & \citet{jiang-bansal-2018-closed} & -0.112\\
Multi-task-Ent-QG & \citet{guo-etal-2018-soft} & -0.117\\
Pointer-Generator &\citet{see-etal-2017-get} & -0.144 \\
UniLM & \citet{UniLM} & -0.151\\
Bottom-Up & \citet{gehrmann-etal-2018-bottom} & -0.160\\
\multicolumn{2}{@{}l}{JEC \quad \quad \quad \quad \quad \quad \quad \quad \citet{xu-durrett-2019-neural}} & -0.167\\
Fast-abs-rl & \citet{chen-bansal-2018-fast} & -0.189\\
NeuralTD & \citet{bohm-etal-2019-better} & -0.215\\
Improve-abs & \citet{kryscinski-etal-2018-improving} & -0.329\\
BertSum-abs & \citet{liu-lapata-2019-text} & -0.341\\
STRASS & \citet{bouscarrat-etal-2019-strass} & -0.405 \\
GPT-2-zero-shot & \citet{gpt2-zero-shot-2019} & -0.441  \\
SENECA & \citet{sharma-etal-2019-entity} & -0.735 \\
\bottomrule[.1em]
\end{tabular}
\caption{CNNDM summarization generators and reference papers.
They are from \citet{fabbri2021summeval}, but we apply detokenization \cite{nltk} and/or truecasing \cite{manning-EtAl:2014:P14-5} to standardize the model outputs for better, reproducible evaluations.
The score column indicates the score from the metric that currently correlates best with the human judgments (\textbf{COMET}).}
\label{tab:cnndm_generators}
\end{table}

\subsection{MSCOCO Image Captioning}
We submit the four strong models from the literature \cite{kasai2021thumb}.\footnote{\url{https://github.com/jungokasai/THumB/tree/master/mscoco}.}
They share similar pipeline structure but vary in model architecture, (pre)training data, model size, and (pre)training objective. 
Table \ref{tab:mscoco_generators} shows the models with their papers and automatic scores from the top-performing metric (\textbf{RefCLIP-S}).

\begin{table}[h!]
\small
\centering
\begin{tabular}{@{} l l c  @{}}
\toprule[.1em]
\textbf{Generator} & \textbf{Description} & \textbf{Score} \\
\midrule[.1em]
VinVL-large\tablefootnote{Model with CIDEr optmization, \url{https://github.com/microsoft/Oscar/blob/master/VinVL_MODEL_ZOO.md\#Image-Captioning-on-COCO}.}& \citet{zhang2021vinvl} & 83.78\\
VinVL-base\tablefootnote{Model with CIDEr optmization.} & \citet{zhang2021vinvl} & 83.45\\
Unified-VLP & \citet{Unified-VLP} & 82.59\\
Up-Down\tablefootnote{Model with cross-entropy optimization, \url{https://vision-explorer.allenai.org/image_captioning}.} & \citet{Anderson2017up-down}& 80.63\\
\bottomrule[.1em]
\end{tabular}
\caption{MSCOCO image captioning generators and reference papers. The score column indicates the score from the metric that currently correlates best with the human judgments (\textbf{RefCLIP-S}).}
\label{tab:mscoco_generators}
\end{table}
\section{Participating Metrics}
\label{appendix:metrics}
Table \ref{tab:metrics} discusses details and configurations of the automatic metrics that we implement in our initial \bilboards.

\begin{table}[h!]
\small
\centering
\begin{tabular}{@{} l @{\hspace{3mm}} l @{} c @{\hspace{1.2mm}} c @{\hspace{1.2mm}} c @{}}
\toprule[.1em]
\textbf{Metric} & \textbf{Description} & \textbf{Refs.} & \textbf{Src.} & \textbf{Cont.} \\
\midrule[.1em]
BLEU\tablefootnote{\textsc{SacreBLEU} implementation of sentence-level BLEU-4; \url{https://github.com/mjpost/sacreBLEU/blob/v1.2.12/sacrebleu.py\#L999}.} & \citet{Papineni2001BleuAM} & \cmark & \xmark&\xmark\\
ROUGE-3\tablefootnote{\url{https://pypi.org/project/rouge-score/}.} & \citet{Lin2004ROUGEAP} & \cmark & \xmark&\xmark\\
ROUGE-L & \citet{Lin2004ROUGEAP} & \cmark & \xmark&\xmark\\
METEOR & \citet{banerjee-lavie-2005-meteor} & \cmark & \xmark & \xmark\\
TER\tablefootnote{\url{https://github.com/mjpost/sacrebleu}.} & \citet{TER2006} & \cmark & \xmark&\xmark\\
METEOR\tablefootnote{\url{https://www.nltk.org/_modules/nltk/translate/meteor_score.html}.} & \citet{banerjee-lavie-2005-meteor} & \cmark & \xmark&\xmark\\
chrF\tablefootnote{\url{https://github.com/m-popovic/chrF}.} & \citet{popovic-2015-chrf} & \cmark & \xmark&\xmark\\
CIDEr\tablefootnote{\url{https://github.com/salaniz/pycocoevalcap}.} & \citet{cider2015} & \cmark & \xmark&\xmark\\
SPICE & \citet{SPICE16} & \cmark & \xmark &\xmark\\
CharacTER\tablefootnote{\url{https://github.com/rwth-i6/CharacTER}.} & \citet{wang-etal-2016-character} & \cmark & \xmark&\xmark\\
chrF++ & \citet{popovic-2017-chrf} & \cmark & \xmark&\xmark\\
SummaQA\tablefootnote{\url{https://github.com/ThomasScialom/summa-qa}.} & \citet{scialom-etal-2019-answers} & \xmark & \cmark&\cmark\\
BERTScore & \citet{bertscore} & \cmark & \xmark & \cmark\\
BLEURT\tablefootnote{\url{https://huggingface.co/metrics/bleurt}.} & \citet{sellam2020bleurt} & \cmark & \xmark & \cmark\\
COMET\tablefootnote{\url{https://github.com/Unbabel/COMET/}.} & \citet{rei-etal-2020-comet} & \cmark & \cmark & \cmark\\
COMET-QE & \citet{rei-etal-2020-comet} & \xmark & \cmark & \cmark\\
Prism-ref\tablefootnote{\url{https://github.com/thompsonb/prism}.} & \citet{thompson-post-2020-automatic} & \cmark & \xmark & \cmark\\
Prism-src & \citet{thompson-post-2020-automatic} & \xmark & \cmark & \cmark\\
CLIP-S\tablefootnote{\url{https://github.com/salaniz/pycocoevalcap}.} & \citet{clipscore} & \xmark & \cmark & \cmark \\
RefCLIP-S & \citet{clipscore} & \cmark & \cmark & \cmark\\
RefOnlyC & \citet{kasai2021thumb} & \cmark & \xmark & \cmark \\
\bottomrule[.1em]
\end{tabular}
\caption{Automatic metrics and their reference papers. The refs., src., and cont.\ columns indicate whether they use references, input source features, and pretrained contextual representations (e.g., BERT; \citealp{devlins2019bert}), respectively.}
\label{tab:metrics}
\end{table}
\section{Additional Ensemble Metric Ablations}
\label{appendix:ensemble-ablations}
Seen in Table \ref{tab:additional-ensemble-ablations} are ablation studies for the ensemble metrics where one of the three selected metrics is removed at a time.
Dropping one metric often has no impact on the correlation score, suggesting that these metrics are highly redundant and capture similar aspects of the output quality. 
\bilboards encourage researchers to explore ways to diversify automatic evaluations by updating the ensemble metric every time a new metric is submitted.

\begin{table}[h]
\centering
\small
\addtolength{\tabcolsep}{-1.5pt}  
\renewcommand{\arraystretch}{1.3}
\begin{tabular}{@{} ccccc @{}}
\toprule
\multirow{2}{*}{\textbf{ZH-EN}}  & -- & COMET & \hlp{COMET-QE} & BLEURT \\
 & 0.61 & 0.61 & 0.57 & 0.61 \\
 \midrule
\multirow{2}{*}{\textbf{EN-DE}}  & -- & COMET & \hlp{COMET-QE} & Prism-ref \\
& 0.51 & 0.52 & 0.52 & 0.52\\
 \midrule
\multirow{2}{*}{\textbf{CNNDM}}  & -- & COMET & \hlp{COMET-QE} & BERTScore \\
& 0.29 & 0.23 & 0.31 & 0.31\\
 \midrule
\multirow{2}{*}{\textbf{COCO}}  & -- & RefCLIP-S & RefOnlyC & CIDEr \\
& 0.45& 0.44& 0.42 & 0.43\\
\bottomrule
\end{tabular}
\caption{Correlations from ensemble ablation studies. One of the three selected metrics is removed at a time, and a new Lasso regression model is trained on the remaining metrics. The bigger the correlation drop is, the bigger the contribution is from the removed metric. \hlp{COMET-QE} is a referenceless metric.}
\label{tab:additional-ensemble-ablations}
\end{table}

\section{Additional Mixed-Effects Analysis}
\label{appendix:mixed-effect}
Table \ref{tab:additional-mixed-effect} presents fixed-effect coefficients that measure how much each automatic metric \textit{overrates} machines over humans (\S\ref{sec:overrate}).
With some exceptions in CNNDM summarization, almost all automatic metrics \textit{underrate} human generations (significantly positive coefficients).
Table \ref{tab:additional-mixed-effect-swapped} swaps the roles of human-generated text, but we still see similar patterns: almost all metrics overrate machines over humans, but the problem is mitigated in COMET-QE, a referenceless, quality estimation metric.
This confirms that our findings hold independently of the design choice.

\begin{table*}[h]
\centering
\small
\addtolength{\tabcolsep}{-2.5pt}  
\renewcommand{\arraystretch}{1.3}
\begin{tabular}{@{} ccccccccc @{}}
\toprule
\multirow{4}{*}{\textbf{ZH-EN}}  & \hlp{COMET-QE} & Ensemble & COMET & BLEURT & BERTScore & CharacTER & MoverScore & METEOR  \\
&\textred{$0.13_{\pm0.01}$} & \textred{$0.26_{\pm0.01}$} & \textred{$0.27_{\pm0.02}$} & \textred{$0.32_{\pm0.02}$}  & \textred{$0.52_{\pm0.02}$}  & \textred{$0.56_{\pm0.02}$}  & \textred{$0.57_{\pm0.02}$}  & \textred{$0.57_{\pm0.02}$}  
\\
& Prism-ref & chrF & TER & chrF++ & ROUGE-3 & BLEU & ROUGE-L & \hlp{Prism-src}\\
& \textred{$0.58_{\pm0.02}$}  & \textred{$0.58_{\pm0.02}$}  & \textred{$0.59_{\pm0.02}$}  & \textred{$0.60_{\pm0.02}$}  & \textred{$0.61_{\pm0.02}$}  & \textred{$0.62_{\pm0.02}$}  & \textred{$0.64_{\pm0.02}$} & \textred{$1.13_{\pm0.02}$}
\\
 \midrule
\multirow{4}{*}{\textbf{EN-DE}} & 
\hlp{COMET-QE} &
Ensemble &
COMET & 
MoverScore &
chrF &
chrF++ &
BLEU &
CharacTER \\
&
\textblue{$-0.17_{\pm0.02}$} &
\textred{$0.03_{\pm0.02}$} &
\textred{$0.08_{\pm0.02}$} &
\textred{$0.22_{\pm0.03}$} &
\textred{$0.29_{\pm0.02}$} & 
\textred{$0.32_{\pm0.02}$} & 
\textred{$0.33_{\pm0.03}$} &
\textred{$0.33_{\pm0.03}$} \\
 
&BERTScore &
Prism-ref &
TER &
\hlp{Prism-src}
&&&&
\\

& \textred{$0.43_{\pm0.02}$} &
\textred{$0.44_{\pm0.02}$} &
\textred{$0.49_{\pm0.03}$} &
\textred{$1.46_{\pm0.03}$} \\

 \midrule
\multirow{4}{*}{\textbf{CNNDM}} &
TER &
COMET &
Ensemble &
BERTScore &
MoverScore &
\hlp{COMET-QE} & 
CharacTER &
BLEURT \\

&\textblue{$-0.58_{\pm0.14}$} &
\textblue{$-0.17_{\pm0.12}$} &
\textblue{$-0.16_{\pm0.12}$} &
$-0.04_{\pm0.12}$ &
$-0.03_{\pm0.11}$ &
$0.02_{\pm0.11}$ &
$0.14_{\pm0.15}$ &
\textred{$0.25_{\pm0.12}$}\\

& \hlp{SummaQA} &
ROUGE-L &
BLEU & 
Prism-ref &
chrF &
chrF++ &
ROUGE-3 &
METEOR\\

&\textred{$0.27_{\pm0.10}$} &
\textred{$0.33_{\pm0.13}$} &
\textred{$0.37_{\pm0.11}$} &
\textred{$0.38_{\pm0.12}$} &
\textred{$0.43_{\pm0.13}$} &
\textred{$0.45_{\pm0.13}$} &
\textred{$0.49_{\pm0.11}$} &
\textred{$0.53_{\pm0.12}$} \\

 \midrule
\multirow{4}{*}{\textbf{COCO}}
&
\hlp{CLIP-S} &
RefCLIP-S &
CharacTER &
chrF &
ROUGE-3 &
chrF++ &
RefOnlyC &
Ensemble 
\\
&
$-0.04_{\pm 0.05}$&
\textred{$0.09_{\pm0.06}$} &
\textred{$0.13_{\pm0.07}$} &
\textred{$0.18_{\pm0.07}$} &
\textred{$0.22_{\pm0.07}$} &
\textred{$0.23_{\pm0.07}$} &
\textred{$0.24_{\pm0.06}$} &
\textred{$0.24_{\pm0.06}$}
\\
&
SPICE &
METEOR &
BLEU &
CIDEr &
ROUGE-L &
BERTScore &
TER &
MoverScore 
\\
&
\textred{$0.25_{\pm0.07}$} &
\textred{$0.32_{\pm0.07}$} &
\textred{$0.39_{\pm0.07}$} &
\textred{$0.43_{\pm0.06}$} &
\textred{$0.44_{\pm0.07}$} &
\textred{$0.45_{\pm0.06}$} &
\textred{$0.45_{\pm0.07}$} &
\textred{$0.51_{\pm0.05}$} 
\\

\bottomrule
\end{tabular}
\caption{Fixed-effect coefficients $\beta_0$ from the linear mixed-effects analysis that measures how much automatic metrics \textbf{overrate} machine text over human, as compared to human raters (\S\ref{sec:overrate}). $\beta_0=0$ is neutral, and statistical significance is indicated by \textred{red} (positive) or \textblue{blue} text (negative). The subscripts indicate 90\% confidence intervals. \hlp{COMET-QE}, \hlp{Prism-src}, \hlp{SummaQA} and \hlp{CLIP-S} are referenceless metrics. In both WMT20 ZH-EN and WMT20 EN-DE, Human-B is evaluated as human-generated translations. Human-A (WMT20 ZH-EN) and Human-A and Human-P (WMT20 EN-DE) are used as the reference set for reference-based metrics.}
\label{tab:additional-mixed-effect}
\end{table*}
\begin{table*}[h]
\centering
\small
\addtolength{\tabcolsep}{-2.5pt}  
\renewcommand{\arraystretch}{1.3}
\begin{tabular}{@{} ccccccccc @{}}
\toprule
\multirow{4}{*}{\textbf{ZH-EN}}  & \hlp{COMET-QE} &
Ensemble &
COMET &
BLEURT &
TER &
BERTScore &
ROUGE-3 &
Prism-ref
\\
&\textred{$0.03_{\pm0.01}$} &
\textred{$0.07_{\pm0.01}$} &
\textred{$0.08_{\pm0.02}$} &
\textred{$0.09_{\pm0.02}$} &
\textred{$0.23_{\pm0.02}$} &
\textred{$0.24_{\pm0.02}$} &
\textred{$0.24_{\pm0.02}$} &
\textred{$0.25_{\pm0.02}$} 
\\
&CharacTER &
ROUGE-L &
chrF &
MoverScore &
METEOR &
chrFpp &
BLEU &
\hlp{Prism-src}
\\
&\textred{$0.25_{\pm0.02}$} &
\textred{$0.26_{\pm0.02}$} &
\textred{$0.27_{\pm0.02}$} &
\textred{$0.27_{\pm0.02}$} &
\textred{$0.29_{\pm0.02}$} &
\textred{$0.29_{\pm0.02}$} &
\textred{$0.30_{\pm0.02}$} &
\textred{$0.79_{\pm0.02}$}

\\
 \midrule
\multirow{4}{*}{\textbf{EN-DE}} & 
\hlp{COMET-QE} &
Ensemble &
COMET &
MoverScore &
Prism-ref &
chrF &
BERTScore &
CharacTER 
\\

&
\textblue{$-0.09_{\pm0.02}$} &
\textblue{$-0.07_{\pm0.02}$} &
\textblue{-$0.06_{\pm0.03}$} &
$0.02_{\pm0.02}$ & 
\textred{$0.18_{\pm0.02}$} &
\textred{$0.20_{\pm0.02}$} &
\textred{$0.21_{\pm0.02}$} &
\textred{$0.22_{\pm0.02}$} 

\\

&
chrF++&
BLEU&
TER &
\hlp{Prism-src}
&&&&
\\

&
\textred{$0.22_{\pm0.02}$} &
\textred{$0.23_{\pm0.02}$} &
\textred{$0.32_{\pm0.02}$} &
\textred{$1.38_{\pm0.03}$} 
&&&&
\\

\bottomrule
\end{tabular}
\caption{Fixed-effect coefficients $\beta_0$ from the linear mixed-effects analysis that measures how much automatic metrics \textbf{overrate} machine text over human, as compared to human raters (\S\ref{sec:overrate}). \textbf{The roles of human translations are swapped}: Human-A is evaluated, and Human-B (WMT20 ZH-EN) and Human-B and Human-P (WMT20 EN-DE) are used as the reference. We still see similar patterns to Table \ref{tab:additional-mixed-effect}: almost all automatic metrics overrate machines over humans, but the problem is less severe in the referenceless metric of \hlp{COMET-QE}.}
\label{tab:additional-mixed-effect-swapped}
\end{table*}

\section{Crowdworker vs.\ Rubric-based Expert Evaluations}
Seen in Table \ref{tab:human-underrate-examples} are examples where crowdworker evaluators \cite{barrault-etal-2020-findings} and professional translators \cite{freitag2021experts} disagree: crowdworkers give lower scores to the human-generated translations than the machine-generated ones.
The first case requires document-level context to properly evaluate.
Document-level context and diversity in high-quality human translations can mislead crowdworkers.
\begin{table*}[h]
\centering
\small
\begin{tabular}{ @{} l | p{7cm}  | p{6cm} @{} }
\toprule
& WMT20 ZH-EN &  \\
 \hline
Source & \begin{CJK}{UTF8}{gbsn} 希望\textbf{兴安省}继续为白俄罗斯企业提供便利条件。 \end{CJK} & \begin{CJK}{UTF8}{gbsn} 凭的是\textbf{相机而动}的时势驾驭。 \end{CJK}    \\
Huoshan & It is hoped that \textbf{Xing'an Province} will continue to provide convenient conditions for Belarusian enterprises.  &  It is based on the current situation of the \textbf{camera}.\\
Human-A &  He hoped that \textbf{Hung Yen Province} would continue to provide convenient conditions for Belarusian enterprises. & This relies on the ability to seize opportunities. \\
Human-B &  He hoped that \textbf{this} could continue in the future. & It is based on the observation of various situations at different times.  \\
\bottomrule
\end{tabular}
\caption{Examples where crowdsource evaluators \cite{barrault-etal-2020-findings} and professional translators \cite{freitag2021experts} disagree: crowdworkers give lower scores to the human-generated translations than the machine-generated ones. The first case requires document-level context to properly evaluate. \begin{CJK}{UTF8}{gbsn}兴安省\end{CJK} is Hung Yen Province in Vietnam in this context, but there is entity ambiguity. (Xing'an Province that existed in the Republic of China.) The second one illustrates the diversity of human generations that misleads crowdworkers.}
\label{tab:human-underrate-examples}
\end{table*}

\end{appendices}

\end{document}